\documentclass[10pt,twocolumn,letterpaper]{article}
\usepackage[pagenumbers]{cvpr}
\input{preamble}
\newif\ifmeasuredmain
\ifdefined\CoWeaveMeasuredEdition\measuredmaintrue\fi
\newcommand{\GoalID}{\pv{80.5}}
\newcommand{\BaseID}{\pv{72.9}}

\newcommand{\GoalC}{\pv{86.5}}
\newcommand{\BaseC}{\pv{77.8}}
\newcommand{\GoalW}{\pv{2.5}}
\newcommand{\BaseW}{\pv{5.9}}
\newcommand{\IndepW}{\pv{9.0}}
\newcommand{\GoalN}{\pv{11.0}}
\newcommand{\BaseN}{\pv{16.3}}
\newcommand{\GoalP}{\pv{82.1}}
\newcommand{\BaseP}{\pv{67.0}}
\newcommand{\DeltaP}{\pv{15.1}}

\newcommand{\Wdelta}{\pv{3.4}}
\newcommand{\Wrelative}{\pv{57.6}}
\newcommand{\WrongMin}{\pv{5.9}}
\newcommand{\WrongMax}{\pv{8.3}}
\newcommand{\TimeDelta}{\pv{8.4}}
\newcommand{\GoalMs}{\pv{46.6}}

\newcommand{\MctrMs}{\pv{49.8}}

\newcommand{\ReaderGain}{\pv{7.1}}
\newcommand{\MatcherGain}{\pv{7.3}}
\newcommand{\BothGain}{\pv{15.8}}

\newcommand{\InteractionPlo}{\pv{$-$1.6}}
\newcommand{\InteractionPhi}{\pv{4.2}}

\newcommand{\TransferMin}{\pv{4.6}}
\newcommand{\TransferMax}{\pv{7.4}}
\newcommand{\TransferPMin}{\pv{9.7}}
\newcommand{\TransferPMax}{\pv{15.6}}
\newcommand{\Arrivals}{\pv{10,783}}
\newcommand{\Complete}{\pv{8,647}}
\newcommand{\Episodes}{\pv{30}}

\newcommand{\BudgetCap}{\pv{32}}

\newcommand{\NativeCo}{\pv{82.1}}
\newcommand{\NativeGmt}{\pv{74.9}}
\newcommand{\NativeDelta}{\pv{7.2}}
\newcommand{\NativeDeltaP}{\pv{14.2}}

\newcommand{\AblInstance}{\pv{8.7}}
\newcommand{\AblFixedRetID}{\pv{1.3}}
\newcommand{\AblInstanceID}{\pv{6.1}}
\newcommand{\PermCost}{\pv{5.1}}
\newcommand{\ZeroCost}{\pv{5.1}}
\newcommand{\RetMarginID}{\pv{7.8}}

\newcommand{\RetMarginP}{\pv{11.7}}

\newcommand{\SweepMinP}{\pv{77.0}}

\newcommand{\RevGranoW}{\pv{3.1}}
\newcommand{\RevGranoP}{\pv{78.4}}
\newcommand{\RevGran}{\pv{3.7}}

\newcommand{\SymArrivals}{\pv{431}}
\newcommand{\SymEpisodes}{\pv{7}}
\newcommand{\TransWcMin}{\pv{3.1}}
\newcommand{\TransWcMax}{\pv{4.0}}
\newcommand{\TransWgMin}{\pv{7.0}}
\newcommand{\TransWgMax}{\pv{7.8}}
\newcommand{\StoreSave}{\pv{6}}
\newcommand{\TransHMin}{\pv{4.9}}
\newcommand{\TransHMax}{\pv{6.8}}

\newcommand{\ReaderMs}{\pv{44.5}}
\newcommand{\TrackletMs}{\pv{45.2}}
\newcommand{\LeadReliable}{\pv{5.9}}
\newcommand{\LeadGroup}{\pv{19.4}}
\newcommand{\LeadSplit}{\pv{24.6}}
\newcommand{\LeadInvalid}{\pv{31.6}}
\newcommand{\QuarterCo}{\pv{74.2}}
\newcommand{\QuarterCoP}{\pv{71.4}}
\newcommand{\DoubleGmt}{\pv{74.2}}
\newcommand{\InvLossCo}{\pv{2.9}}
\newcommand{\InvLossGmt}{\pv{7.1}}
\newcommand{\SdMax}{\pv{0.8}}
\def\RowsMainwide{%
Independent & \pv{68.8} & \pv{57.5} & \pv{68.4} & \pv{9.0} & \pv{22.6} & \pv{55.3} &  & \pv{40.2} & \pv{1.00} &  & \pv{65.0} & \pv{50.7} &  & \pv{61.9} & \pv{47.7} &  & \pv{58.5} & \pv{44.6} \\
DyGLIP & \pv{71.3} & \pv{59.0} & \pv{70.3} & \pv{8.3} & \pv{21.4} & \pv{58.6} &  & \pv{39.5} & \pv{0.99} &  & \pv{67.9} & \pv{54.3} &  & \pv{63.9} & \pv{51.6} &  & \pv{61.0} & \pv{47.2} \\
ReST & \pv{72.1} & \pv{60.9} & \pv{72.7} & \pv{7.6} & \pv{19.8} & \pv{62.2} &  & \pv{37.5} & \pv{0.88} &  & \pv{68.4} & \pv{59.1} &  & \pv{65.8} & \pv{55.4} &  & \pv{61.8} & \pv{51.6} \\
MCTR & \pv{72.7} & \tsecond{\pv{62.7}} & \pv{74.5} & \pv{6.5} & \pv{18.9} & \pv{64.2} &  & \pv{49.8} & \pv{0.71} &  & \pv{69.1} & \pv{58.9} &  & \pv{66.1} & \pv{56.2} &  & \pv{62.6} & \pv{54.6} \\
GMT (causal) & \tsecond{\pv{72.9}} & \pv{62.6} & \tsecond{\pv{77.8}} & \tsecond{\pv{5.9}} & \tsecond{\pv{16.3}} & \tsecond{\pv{67.0}} &  & \pv{43.0} & \pv{1.00} &  & \tsecond{\pv{70.2}} & \tsecond{\pv{63.5}} &  & \tsecond{\pv{66.6}} & \tsecond{\pv{59.9}} &  & \tsecond{\pv{63.1}} & \tsecond{\pv{56.0}} \\
\rowcolor{tabours}\textbf{CoWeave} & \tbest{\pv{80.5}} & \tbest{\pv{69.3}} & \tbest{\pv{86.5}} & \tbest{\pv{2.5}} & \tbest{\pv{11.0}} & \tbest{\pv{82.1}} &  & \pv{46.6} & \pv{0.94} &  & \tbest{\pv{74.8}} & \tbest{\pv{73.2}} &  & \tbest{\pv{74.0}} & \tbest{\pv{75.5}} &  & \tbest{\pv{69.3}} & \tbest{\pv{68.3}} \\
$\Delta$ & \dgood{\pv{$+$7.6}}\dci{\pv{1.1}} & \dgood{\pv{$+$6.7}} & \dgood{\pv{$+$8.7}}\dci{\pv{1.5}} & \dgood{\pv{$-$3.4}}\dci{\pv{0.3}} & \dgood{\pv{$-$5.3}} & \dgood{\pv{$+$15.1}}\dci{\pv{1.8}} &  & \tna & \tna &  & \dgood{\pv{$+$4.6}}\dci{\pv{1.1}} & \dgood{\pv{$+$9.7}} &  & \dgood{\pv{$+$7.4}}\dci{\pv{0.5}} & \dgood{\pv{$+$15.6}} &  & \dgood{\pv{$+$6.2}}\dci{\pv{1.5}} & \dgood{\pv{$+$12.3}} \\%
}
\def\RowsFrontend{%
Independent & \pv{70.1} & \pv{57.1} & \pv{1.00} \\
DyGLIP & \pv{73.1} & \pv{61.0} & \pv{1.17} \\
ReST & \pv{73.7} & \pv{64.3} & \pv{1.06} \\
MCTR & \pv{74.1} & \pv{66.2} & \pv{1.39} \\
GMT (causal) & \tsecond{\pv{74.9}} & \tsecond{\pv{69.8}} & \pv{1.26} \\
\rowcolor{tabours}\textbf{CoWeave} & \tbest{\pv{82.1}} & \tbest{\pv{84.0}} & \pv{1.34} \\
$\Delta$ & \dgood{\pv{$+$7.2}} & \dgood{\pv{$+$14.2}} &  \\%
}
\def\RowsAblationmain{%
\rowcolor{tabgroup}\multicolumn{5}{l}{\textit{\textbf{Single-part replacement, same read budget}}} \\
\rowcolor{tabours}\textbf{Full CoWeave} & \tbest{\pv{80.5}} & \tbest{\pv{2.5}} & \tbest{\pv{82.1}} & \tna \\
Fixed retrieval & \pv{79.2} & \pv{3.3} & \pv{77.8} & \dbad{\pv{$-$4.3}} \\
Zero relational support & \pv{77.1} & \pv{3.5} & \pv{77.0} & \dbad{\pv{$-$5.1}} \\
Tracklet-level reassignment & \pv{78.6} & \pv{3.1} & \pv{78.4} & \dbad{\pv{$-$3.7}} \\
Fixed record ownership & \pv{76.5} & \pv{3.8} & \pv{74.9} & \dbad{\pv{$-$7.2}} \\
Without instance detail & \pv{74.4} & \pv{4.5} & \pv{73.4} & \dbad{\pv{$-$8.7}} \\
\rowcolor{tabgroup}\multicolumn{5}{l}{\textit{\textbf{Retention policy, same storage cap}}} \\
\rowcolor{tabours}\textbf{Full CoWeave} & \tbest{\pv{80.5}} & \tbest{\pv{2.5}} & \tbest{\pv{82.1}} & \tna \\
Recent & \pv{72.1} & \pv{6.9} & \pv{65.8} & \dbad{\pv{$-$16.3}} \\
Diverse & \pv{72.7} & \pv{6.5} & \pv{68.4} & \dbad{\pv{$-$13.7}} \\
Task-aware independent & \pv{72.1} & \pv{5.8} & \pv{70.4} & \dbad{\pv{$-$11.7}} \\%
}

\newcommand{\RepairEver}{\pv{75}}
\newcommand{\RepairMedian}{\pv{5}}
\newcommand{\RepairHeld}{\pv{69}}

\definecolor{cvprblue}{rgb}{0.21,0.49,0.74}
\usepackage[pagebackref,breaklinks,colorlinks,allcolors=cvprblue]{hyperref}
\hypersetup{pdfauthor={},pdftitle={CoWeave: Identity-Specific Context Reasoning for Persistent Multi-Camera Tracking}}
\def\confName{CVPR}
\def\confYear{2027}
\title{\raisebox{-0.15em}{\includegraphics[height=1.30em]{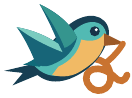}}\hspace{0.24em}CoWeave: Identity-Specific Context Reasoning\\for Persistent Multi-Camera Tracking}
\author{Haoyang Wu\textsuperscript{1,*}\hspace{0.75em}Taoran Fu\textsuperscript{2,*}\hspace{0.75em}Shoudong Han\textsuperscript{1,\textdagger}\hspace{0.75em}Chaoyue Li\textsuperscript{1}\hspace{0.75em}Ziqian Wang\textsuperscript{3}\hspace{0.75em}Jiachen Luo\textsuperscript{4}\\
Zhenhao Zhang\textsuperscript{5}\hspace{0.75em}Sijia Chen\textsuperscript{1}\hspace{0.75em}Heng Li\textsuperscript{1}\hspace{0.75em}Hao Geng\textsuperscript{6}\hspace{0.75em}Liuzhou Zhang\textsuperscript{7}\hspace{0.75em}Zhenyang Xie\textsuperscript{8}\\
{\small \textsuperscript{1}Huazhong University of Science and Technology\quad \textsuperscript{2}Xi'an Jiaotong-Liverpool University}\\
{\small \textsuperscript{3}Beijing Institute of Technology\quad \textsuperscript{4}Independent Researcher}\\
{\small \textsuperscript{5}Tsinghua University\quad \textsuperscript{6}Shanghai Maritime University}\\
{\small \textsuperscript{7}The Hong Kong University of Science and Technology\quad \textsuperscript{8}Jiangxi University of Water Resources and Electric Power}\\[-1pt]{\small \textsuperscript{*}Equal contribution.\qquad \textsuperscript{\textdagger}Corresponding author.}}
\hypersetup{pdfauthor={Haoyang Wu, Taoran Fu, Shoudong Han, Chaoyue Li, Ziqian Wang, Jiachen Luo, Zhenhao Zhang, Sijia Chen, Heng Li, Hao Geng, Liuzhou Zhang, Zhenyang Xie},pdftitle={CoWeave: Identity-Specific Context Reasoning for Persistent Multi-Camera Tracking}}
\begin{document}
\maketitle

\begin{abstract}
\begingroup\parfillskip=0pt
Multi-camera tracking must associate each cross-camera arrival with an identity and preserve that identity in memory. Two coupled failures undermine this process: shared context can support competing identity assignments equally, and incorrect memory attribution can propagate errors to later arrivals. We propose \textbf{CoWeave}, which unifies assignment scoring, evidence retrieval, and memory maintenance through a common identity decision loss. Candidate-conditioned scoring compares joint assignments using evidence that distinguishes them and allows unresolved outputs when committing is too costly. A budgeted reader selects historical sources by their predicted reduction in decision loss, sharing each source across active people without duplicate access. Memory stores each observation with revisable identity ownership and separately tracks the validity of its relations. On MTMMC, under a common causal stream and budget, CoWeave improves global identification F1 from 72.9 to 80.5 over causal GMT and reduces wrong arrivals from 5.9\% to 2.5\%. Among arrivals with 30 subsequent observations, the fraction remaining correct from arrival throughout follow-up rises from 67.0\% to 82.1\%. Gains also hold with each method's native front end and on WILDTRACK, MMPTrack, and DIVOTrack.\par\endgroup

\end{abstract}

\begin{figure}[t]
\begin{gridfloat}
\centering
\input{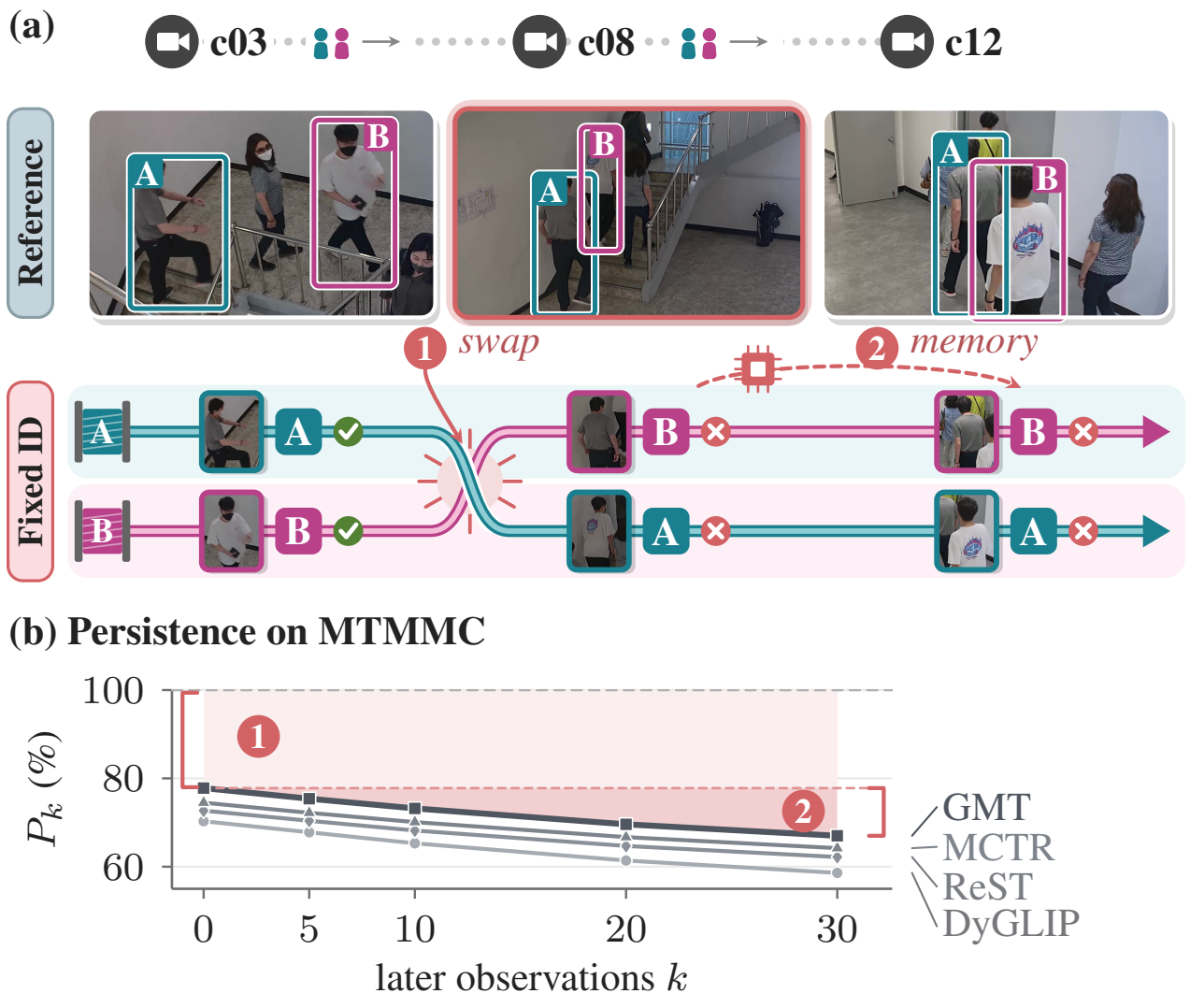}
\caption{\textbf{Identity swaps at cross-camera arrivals persist in memory.} (a) Two people in MTMMC s16, GT~113 as A and GT~026 as B, pass cameras c03, c08, and c12. Lane tint marks the true person and yarn the identity that a tracker with fixed identities writes. The yarns cross where the two arrive side by side at c08 (1), and memory carries the swap to c12 (2). (b) For all four association pipelines, the share $P_k$ of arrivals with $k$ later observations that stay correct throughout lies below 78\% at arrival (1) and keeps falling over subsequent observations as $k$ grows (2).}
\label{fig:motivation}
\end{gridfloat}
\end{figure}

\begin{figure*}[t]
\begin{gridfloat}
\centering
\input{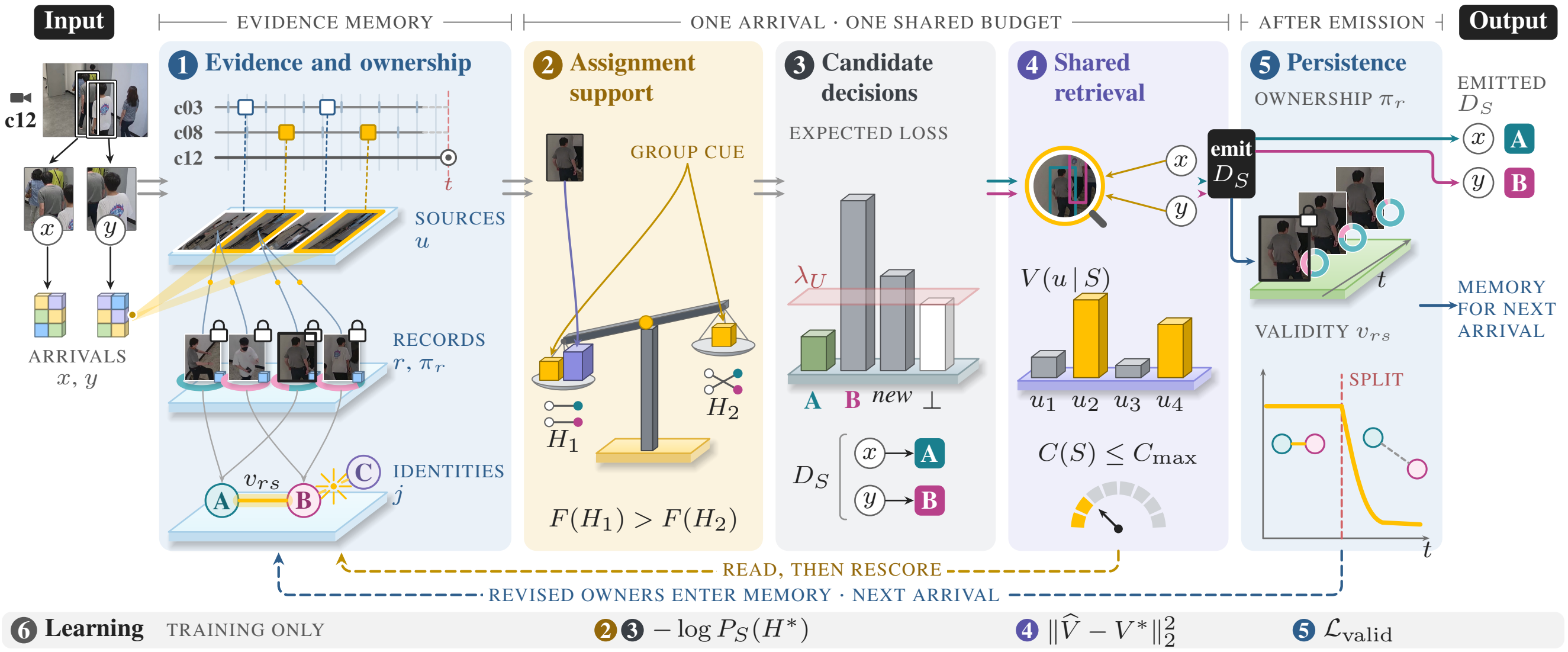}
\caption{\textbf{CoWeave at one arrival.} (1) Memory keeps sources $u$, frames tied from each camera stream before the current time $t$ (gold: read at this arrival), and records $r$ whose pixels stay locked while their ownership $\pi_r$ can change (ring colours: shares of A and B); $v_{rs}$ is the validity of a relation between records, drawn between the identities they support, and C is another known identity. (2) The group cue (gold) loads $H_1$ and $H_2$ equally, whereas the outlined record of A (lavender) loads only $H_1$. (3) The candidate $D_S$ has the lowest expected loss, here below the cost $\lambda_U$ of leaving $x$ unresolved. (4) Reads by value $V(u\mid S)$ under one budget rescore (2) and (3), one read of $u_2$ serves $x$ and $y$, and $D_S$ is emitted once no read has positive value. (5) After emission, $\pi_r$ moves to A and $v_{rs}$ drops after a split. (6)~The training objective supervises assignment scoring, relation validity, and the value of historical reads in (2)--(5).}
\label{fig:overview}
\end{gridfloat}
\end{figure*}

\section{Introduction}
\label{sec:introduction}
Multi-camera tracking takes synchronized video streams from several cameras and outputs, for every visible person, one identity that persists across viewpoint changes, occlusions, and illumination differences~\cite{ristani2016performance,chavdarova2018wildtrack,han2023mmptrack,woo2024mtmmc}. An association stage links local tracks, built from appearance and motion cues~\cite{bewley2016sort,wojke2017deep,ristani2018features,zhang2022bytetrack}, across cameras. At each cross-camera \emph{arrival}, where a person reappears after being observed in another view, this stage decides, under a limited read budget, which established identity the person continues and which observations enter that identity's memory. Recent association methods support these decisions with context and memory, through relation graphs and attention over tracklets~\cite{quach2021dyglip,li2021visio,nguyen2022lmgp,luna2023gnncca,cheng2023rest}, long-term memory attention~\cite{cai2022memot,gao2023memotr}, probabilistic view-to-global association in the Multi Camera Tracking Transformer (MCTR)~\cite{niculescumizil2025mctr}, and cross-view histories, neighboring-target relations, and occlusion memory in GMT~\cite{zhen2026gmt}. Context lets a tracker draw on the people around a target when appearance alone is ambiguous, and memory lets a tracker compare an arrival against earlier observations from every view.

These methods are evaluated by identity measures and Higher Order Tracking Accuracy (HOTA) computed over whole sequences~\cite{ristani2016performance,luiten2021hota}, so neither the decision at an arrival nor the memory it leaves behind is scored on its own. Figure~\ref{fig:motivation}(a) shows a case that such scores hide. Two similarly dressed people walk together from camera c03 to c08 and c12. At c08 they arrive side by side, seen from behind, and a tracker that writes each match as a fixed identity swaps them. The swapped observations enter memory, and the swap reappears at c12. The failure is not limited to this pair. On MTMMC~\cite{woo2024mtmmc}, four published association pipelines, the dynamic graph model with link prediction (DyGLIP)~\cite{quach2021dyglip}, the reconfigurable spatial-temporal graph model (ReST)~\cite{cheng2023rest}, MCTR, and GMT, give an established identity to the wrong person at \WrongMin\% to \WrongMax\% of arrivals. Figure~\ref{fig:motivation}(b) shows that even GMT, run causally, keeps only \BaseP\% of the arrivals with 30 later observations correct throughout every subsequent observation.

An arrival can go wrong at two points, in the score that compares competing assignments and in the memory that the decision leaves behind. In the score, the co-occurrence of two people who walk together shows that they form a group but not who is who. A context bonus that depends only on the arriving person or only on the candidate identity sums to the same value under both pairings, so it adds the same support to the correct assignment and to the swapped one, however much is read, and a similarity ranking spends the budget on familiar group views that leave the pairings tied. Relational trackers such as ReST and GMT~\cite{cheng2023rest,zhen2026gmt} encode neighbors into per-observation features or pairwise affinities before assignment, so a group cue shared by both people raises both pairings alike, and a stronger descriptor~\cite{zhou2019osnet,luo2019bag,he2021transreid} still leaves two similarly dressed people close from behind. In memory, the result of one arrival becomes evidence for the next. Memory-based trackers such as MeMOTR and GMT~\cite{gao2023memotr,zhen2026gmt} fold matched observations into a per-identity memory and provide no operation that moves a stored observation to another identity, and corrective matching as in OCMCTrack~\cite{specker2024ocmctrack} reassigns whole tracks rather than single observations. If an image of person B is written into the memory of person A, a later view of B resembles the history of A and can be matched to A again. Stored relations also go stale once a group separates. These failures point to three targets: comparing competing assignments with evidence specific to each, spending reads only on what can change this comparison, and keeping both the attribution of stored observations and the links between observations open to revision at any time.

Based on this analysis, we propose \textbf{CoWeave}, which connects scoring, retrieval, and memory through one identity decision loss, the expected cost of wrong, unmatched new, and unresolved outputs, as Figure~\ref{fig:overview} illustrates. To compare competing assignments with evidence specific to each, CoWeave scores every joint assignment with candidate-conditioned instance and relational evidence, so that context counts as identity-specific evidence only when it supports one pairing more than the other, and it leaves an arrival unresolved when no evidence separates the candidates. To spend reads only on what can change this comparison, it reads and retains historical sources, which are stored frames or observation bundles, by their expected reduction of the decision loss. All active people share one read budget, and a source that serves several of them is charged once. To keep attribution and links open to revision, it stores each observation once with its \emph{ownership}, a distribution over candidate identities that later evidence can revise, and tracks the \emph{relation validity} between two observations, whether their link still holds, apart from ownership.

We compare CoWeave with five pipelines on MTMMC, the four published trackers above and a non-relational baseline, under a common stream and budget. CoWeave keeps \GoalP\% of the arrivals that have 30 later observations correct throughout, against \BaseP\% for GMT, and raises the global identification F1 score (G-IDF1) from \BaseID{} to \GoalID{}. With detections, descriptors, and candidates held fixed, adopting only its source selection or only its assignment scoring raises this persistence by \ReaderGain{} or \MatcherGain{} points. In a paired replay of 100 ownership errors, revisable ownership holds the original identity at the 30th later observation in \RepairHeld{} cases and frozen ownership in none, and revising single records keeps \RevGran{} points more arrivals correct throughout than reassigning whole tracklets. The lead remains when every pipeline runs its own front end, and on WILDTRACK, MMPTrack, and DIVOTrack CoWeave leads GMT by \TransferMin{} to \TransferMax{} G-IDF1 points. Our contributions are as follows:
\begin{itemize}
\item \textbf{Two failure modes and an arrival-level evaluation.} We identify context that supports competing assignments equally and wrong matches that memory keeps, and we score emitted errors and uninterrupted persistence at each eligible cross-camera arrival in the test stream.
\item \textbf{One decision loss for scoring, retrieval, and memory.} CoWeave reads addressable sources shared by all active people, keeps the ownership of each stored observation apart from the validity of its relations, and scores, reads, and retains evidence by the expected cost of the identity decisions it emits under the shared read budget.
\item \textbf{Controlled evidence on four datasets.} We compare CoWeave with five pipelines under shared and native front ends and test each of its modules with matched controls, including a paired replay of ownership errors and a tracklet-level revision control.
\end{itemize}

\section{Related Work}
\label{sec:related}

\runin{Within-camera tracking.}Tracking-by-detection links detections through motion, appearance, or box regression~\cite{bewley2016sort,wojke2017deep,ristani2018features,bergmann2019tracktor}, message passing over detection graphs~\cite{braso2020neural}, joint detection and embedding~\cite{zhang2021fairmot,pang2021quasidense}, trajectory queries~\cite{meinhardt2022trackformer,zeng2022motr,zhou2022global}, and association of low-confidence detections or observations~\cite{zhang2022bytetrack,cao2023ocsort,seidenschwarz2023ghost}. In these designs, temporal windows and representation updates determine what information reaches a later cross-camera arrival. Unlike trackers that carry identity in queries or updated embeddings, CoWeave keeps each historical frame or observation bundle as an addressable source whose content is never rewritten.

\runin{Relational association.}Tracklet relations and graphs link observations across views~\cite{he2020tracta,kohl2020mta,quach2021dyglip,luna2023gnncca}. Early methods associate people across tracklets~\cite{he2020tracta,kohl2020mta}, use symmetry and transitivity~\cite{gan2021mvmhat}, predict dynamic tracklet links~\cite{quach2021dyglip}, or condition tracklets on other candidates~\cite{li2021visio}. Recent works combine geometry with lifted multicuts~\cite{nguyen2022lmgp}, pass messages across views~\cite{luna2023gnncca}, reconfigure spatial and temporal graphs online~\cite{cheng2023rest}, or use neighboring people alongside RGB--thermal fusion~\cite{fan2025allday}. Their context enters as pairwise affinities or as features fixed before assignment, so a cue shared by two group members raises both of their pairings alike, and online methods can carry an accepted handover error into subsequent cross-camera matching decisions.

\runin{Closest trackers.}GMT~\cite{zhen2026gmt} uses cross-view histories, neighboring-target relations, and occlusion memory, MCTR~\cite{niculescumizil2025mctr} uses view-specific cross-attention and probabilistic association, and OCMCTrack~\cite{specker2024ocmctrack} detaches inconsistent single-camera tracks and rematches lost ones. All three keep identity state per track, so none can reassign an individual stored observation, and the GMT authors leave errors within trajectories to future work. MCTR relates people only through self-attention among track queries, OCMCTrack not at all, and the relation cues in the trajectory features of GMT may persist after the people separate. CoWeave instead attributes each stored observation through a revisable ownership distribution, tracks relation validity separately, counts context only when it changes the relative scores of competing assignments, and selects sources by their expected reduction of the decision loss under one budget for all active people. Multiple hypothesis tracking~\cite{reid1979mht} and joint probabilistic data association~\cite{fortmann1983jpda} also weigh competing assignments but settle them before a decision is final, whereas CoWeave revises the owner of single stored observations after emission, which changes the evidence for later arrivals but not emitted identities.

\runin{Memory and selective retrieval.}Both strengthen instance evidence and limit what a tracker has to search. Re-identification work adds part pooling~\cite{sun2018pcb}, visible-part matching~\cite{gao2020pose}, multiscale features~\cite{zhou2019osnet}, stronger training~\cite{luo2019bag}, view-conditioned transformers~\cite{he2021transreid}, and distractor-suppressing video memory~\cite{eom2021memory}. Recent trackers reference spatiotemporal memory~\cite{cai2022memot} or long-term memory attention~\cite{gao2023memotr}, and video segmentation shares correspondence and consolidates memory~\cite{oh2019stm,cheng2021stcn,cheng2022xmem,zhou2024rmem}. Camera-network systems prune searches through spatial and temporal correlations~\cite{jain2020spatula} or adapt query processing under a recall constraint~\cite{chunduri2025tracer}. Both systems search for one query identity, whereas at an arrival several people compete for the same sources. CoWeave selects shared sources for all active people by expected loss reduction after prior reads.

\runin{Benchmarks and metrics.}Persistent tracking is scored with CLEAR, identity measures, and HOTA~\cite{bernardin2008clear,ristani2016performance,luiten2021hota}. Earlier benchmarks cover calibrated pedestrian views~\cite{chavdarova2018wildtrack}, vehicles~\cite{tang2019cityflow}, and within-camera tracking~\cite{dendorfer2021motchallenge,sun2022dancetrack}, and recent cross-camera datasets add several environments~\cite{han2023mmptrack}, open scenes~\cite{hao2024divotrack}, and long synchronized RGB and thermal streams~\cite{woo2024mtmmc}. Sequence-level scores, however, do not isolate what happens at an arrival or in the observations that follow it. We therefore score continuation and persistence at every eligible arrival and across its later observations.

\section{CoWeave}
\label{sec:method}

CoWeave scores every joint assignment of the current observations from the sources read so far, reads further sources by their expected reduction of the identity decision loss under one shared budget, emits the lowest-loss decision, and writes records whose ownership later evidence can revise, as Figure~\ref{fig:overview} illustrates for one arrival.

\subsection{Evidence and ownership}
\label{sec:evidence}

Reading selectively and revising later both need stored content that is not welded to one identity. At decision time $t$, $X_t=\{x_i\}_{i=1}^{n_t}$ denotes current observations and $\mathcal J_t$ the known identities, where each $x_i$ holds a camera index, time interval, descriptor, and local track cues computed without identity ground truth. A source $u$ is an addressable historical frame or observation bundle with a unique provenance key. Because availability time $a_u$ includes processing, only sources with $a_u\leq t$ are eligible.

Ownership attaches to instance records because one source can contain several people. Each record $r$ keeps its source key, local observation key, descriptor $z_r$, and distribution $\pi_r$ over known identities and an unknown state $\varnothing$, and the source content stays unchanged when $\pi_r$ changes. Context records keep co-occurrence, supporting sources, and the relation validity $v_{rs}(t)\in[0,1]$ for each pair of historical observation records denoted by $r$ and $s$.

The memory $\mathcal M_t$ comprises a source bank for evidence, instance records for possible owners, and a context index for relations guiding retrieval, as Figure~\ref{fig:overview}(1) illustrates.

\subsection{Assignment support}
\label{sec:relative_support}

The records in $\mathcal M_t$ become identity evidence only through the assignments they support. Let $H$ assign observations to existing or new identities with $j_i=H(i)$, whereas an emitted decision $D$ may also leave observations unresolved. Reading a source exposes records $\mathcal R_u$ and relations, whose support for an assignment $H$ is
\begin{equation}\gridfix{-5.16pt}
\label{eq:source_support}
\begin{split}
\Phi_u(H) ={}& \sum_i \sum_{r\in\mathcal R_u}
  w_{ir}(j_i) f_\theta(x_i,z_r) \\
 &+\lambda_{\mathrm{rel}}\sum_{i<\ell}\sum_{\substack{r,s\in\mathcal R_u\\r\ne s}}
 w_{ir}(j_i)w_{\ell s}(j_\ell)\kappa_{rs}(j_i,j_\ell) \\
 &\hspace{10mm}\cdot v_{rs}(t)g_\theta(x_i,x_\ell,z_r,z_s).
\end{split}
\end{equation}
Here $w_{ir}(j)=\alpha_{ir}\pi_r(j)$ weights record $r$ by its ownership under the proposed identity $j$, with attention $\alpha_{ir}$ normalized within $\mathcal R_u$, and $\lambda_{\mathrm{rel}}\geq0$ weights the relational term. The mask $\kappa_{rs}(j,k)$ is zero only when $H$ gives $x_i$ and $x_\ell$ the same identity, $j=k$, while $r$ and $s$ show different people, such as two detections in one frame, so relations between two people under $j\neq k$ are kept. Because $w_{ir}(j)$ scales a record by its ownership under $j$, one source can support an assignment without supporting its permutation. The functions $f_\theta$ and $g_\theta$ compare instances and relations under $H$ with camera and time descriptors. For fresh labels, $w_{ir}(j)=0$ and separate birth scores apply.

Summing this support over a set $S$ of sources already read gives the joint score of an assignment
\begin{equation}\gridfix{0.6pt}
\label{eq:score}
F(H;S)=\sum_i b_i\bigl(H(i)\bigr)
       +\sum_{u\in S}\Phi_u(H).
\end{equation}
The base score $b_i$ uses the observation and each candidate's resident summary. As provenance tracks feature use, each source enters once with detail absent from that summary.

For the two-person ambiguity between $H_1=(x\mapsto A,y\mapsto B)$ and $H_2=(x\mapsto B,y\mapsto A)$, a source can resolve the competition only by changing the evidence contrast $\Delta(S)=F(H_1;S)-F(H_2;S)$. A term $a_i$ that depends only on the observation, or a term $d_j$ that depends only on the identity, cannot change $\Delta$, because its sum is the same for every permutation of the same matched sets. New-identity assignments and unresolved decisions change the matched sets and lie outside this invariance. To separate permutations, Equation~\ref{eq:source_support} exposes assignment-dependent instance pairs, whereas symmetric observations and relations, like the group cue on both pans in Figure~\ref{fig:overview}(2), leave the assignments indistinguishable. The contrast ranks complete assignments but fixes neither the feasible assignment set nor the decision to abstain.

\subsection{Identity decisions}
\label{sec:decision}

The joint score $F(H;S)$ is compared only among feasible assignments, which joint inference groups into connected ambiguity components linked by shared historical candidates, admissible fresh-group links, observation incompatibilities, and active relations. A fresh group is a set of observations that share one new label. Given retained candidates $\mathcal C_i\subseteq\mathcal J_t$, fresh labels $\mathcal B_t$, and incompatible pairs $\mathcal I_t$, including simultaneous same-camera detections after duplicates, the feasible assignment set can be written as
\begin{equation}
\label{eq:feasible}
\begin{split}
\mathcal H_t=\{H:\;&H(i)\in\mathcal C_i\cup\mathcal B_t,\\
 &H(i)\ne H(\ell),\ \forall(i,\ell)\in\mathcal I_t\}.
\end{split}
\end{equation}
Compatible cross-camera observations may then share an identity, and fresh-group links keep each admissible group within one component so the relabeling loss separates across components. Naming each fresh group by its smallest observation index prevents duplicate hypotheses.

We define $P_S(H)\propto\exp(F(H;S)/T)$ over the retained hypotheses in $\mathcal H_t$ with calibration temperature $T>0$~\cite{guo2017calibration}. The feasible decision $D_S$ minimizes the expected identity decision loss $\sum_HP_S(H)L(D,H)$, where each output $d_i=D(i)$ is an existing identity, a fresh label, or an unresolved output $\bot$, and the identity decision loss is defined as
\begin{equation}\gridfix{1.43pt}
\label{eq:loss}
L(D,H)=\min_{\sigma}\sum_i\ell_\sigma\bigl(d_i,H(i)\bigr).
\end{equation}
Here $\sigma$ matches fresh groups in $D$ to those in $H$ one-to-one, so new identities are compared up to relabeling, and $\ell_\sigma$ charges $\lambda_W=1$ for a wrong established ID, $\lambda_B=0.5$ for an unmatched fresh label, $\lambda_U=0.35$ for an unresolved output, which acts as an abstention~\cite{geifman2019selectivenet}, and zero otherwise. For a single observation without coupling to others, leaving it unresolved therefore costs less than emitting an established identity whenever that identity's posterior probability is below $1-\lambda_U/\lambda_W$, which includes every posterior tie between two candidate identities.

Finite candidate, hypothesis, and decision beams approximate this minimization, and a decision beam ranks partial decisions by posterior expected loss before emitting its lowest-risk completed decision. As $D_S$ is defined for any read set, the same loss prices every unread source.

\subsection{Shared retrieval}
\label{sec:retrieval}

A source is worth reading only if it can lower the decision loss of Eq.~\ref{eq:loss}. Given $\mathcal F_S$, comprising current observations, resident summaries, accessible metadata, and already-read content without unread payloads, we value source $u$ by the expected loss reduction of the decision rule
\begin{equation}\gridfix{-3.41pt}
\label{eq:read_value}
\begin{split}
V(u\mid S)=\mathbb E\bigl[&L(D_S,H)\\
 &-L(D_{S\cup\{u\}},H)\mid\mathcal F_S\bigr].
\end{split}
\end{equation}
Although this expectation covers unknown identities and unread content, its predictor uses only $\mathcal F_S$ and metadata. Read value thus measures how much a read is expected to improve the decision, whereas a similarity ranking favors sources that resemble the query even when they cannot change it. It matches the regression target below and equals Bayes-risk reduction when $P_S$ is the true conditional posterior and decisions are exact. A source that raises both competing assignments or leaves them tied has zero value.

Sources are selected greedily by predicted value per incremental cost, and each read updates the assignment distribution and the remaining value predictions, so conditioning on $S$ accounts for evidence already read. One source may serve several people at a single read cost, and the shared budget bounds the total cost of reading and selection actions
\begin{equation}\gridfix{4.54pt}
\label{eq:read_budget}
C(S)=\sum_{u\in S}c_{\mathrm{read}}(u)
      +\sum_{k=1}^{K}c_{\mathrm{select},k}\leq C_{\max}.
\end{equation}
Here the second term charges the $K$ selection actions, including index expansion and reassessment, in one unit, and every action must fit the remaining budget $C_{\max}-C(S)$. Retrieval stops once no single read has positive predicted value, the budget runs out, or the deadline passes.

\begin{figure}[t]
\begin{gridfloat}
\centering
\begin{tikzpicture}[cw,font=\footnotesize]
\path (0,0) (237,110);
\cwpcard{0}{50}{237}{110}{(a) 100 replayed ownership errors at $h{=}30$}
\cwpcard{0}{0}{237}{46}{(b) Repairs and new breaks}
\def\BW{1.66}
\foreach \x/\c/\t in {10/cwOurs/original identity,84/cwBad!75/wrong identity,150/cwB4/new or unresolved}{
  \path[fill=\c,rounded corners=1pt] (\x,90.6) rectangle ++(6,4.6); \node[anchor=base west,inner sep=1pt] at (\x+6.5,90.6) {\t};}
\foreach \y/\name/\o/\w/\n in {78/Frozen/0/78/22,64/Revisable/69/15/16}{
  \node[anchor=east,inner sep=0pt] at (47,\y) {\name};
  \pgfmathsetmacro{\xa}{51}\pgfmathsetmacro{\xb}{51+\o*\BW}\pgfmathsetmacro{\xc}{\xb+\w*\BW}\pgfmathsetmacro{\xd}{\xc+\n*\BW}
  \begin{scope}\clip[rounded corners=3pt] (\xa,\y-5.5) rectangle (\xd,\y+5.5);
  \ifnum\o>0 \fill[cwOurs] (\xa,\y-5.5) rectangle (\xb,\y+5.5);\fi
  \fill[cwBad!75] (\xb,\y-5.5) rectangle (\xc,\y+5.5);
  \fill[cwB4] (\xc,\y-5.5) rectangle (\xd,\y+5.5);
  \fill[white,opacity=0.18] (\xa,\y+1.5) rectangle (\xd,\y+5.5);
  \end{scope}
  \ifnum\o>0 \node[text=white,font=\footnotesize\bfseries] at ({(\xa+\xb)/2},\y) {\o};\fi
  \node[text=white] at ({(\xb+\xc)/2},\y) {\w};
  \node[text=cwInk] at ({(\xc+\xd)/2},\y) {\n};
  \ifnum\o>0 \draw[white,line width=0.8pt] (\xb,\y-5.5) -- (\xb,\y+5.5);\fi \draw[white,line width=0.8pt] (\xc,\y-5.5) -- (\xc,\y+5.5);}
\def\SC{0.85}
\node[sc,anchor=west,text=cwOurs] at (52,27.5) {errors repaired};
\node[sc,anchor=west,text=cwBad!85!black] at (156,27.5) {correct ones broken};
\draw[cwInk!40,line width=0.4pt] (150,3) -- (150,31);
\node[anchor=east,inner sep=0pt] at (47,16) {Revisable};
\node[anchor=east,inner sep=0pt] at (47,6.5) {Frozen};
\draw[cwInk!60,line width=0.5pt] (52,2.5) -- (52,20.5);
\path[fill=cwOurs,rounded corners=1.5pt] (52,13) rectangle ++(69*\SC,6);
\node[anchor=west,text=cwOurs,font=\footnotesize\bfseries,inner sep=1.5pt] at (52+69*\SC,16) {69/100};
\node[anchor=west,text=cwNote,inner sep=1.5pt] at (52,6.5) {0/100};
\draw[cwInk!60,line width=0.5pt] (156,2.5) -- (156,20.5);
\path[fill=cwBad!75,rounded corners=0.5pt] (156,13) rectangle ++(1*\SC,6);
\node[anchor=west,text=cwBad!85!black,font=\footnotesize\bfseries,inner sep=1.5pt] at (156+1*\SC+1,16) {2/200};
\node[anchor=west,text=cwNote,inner sep=1.5pt] at (156,6.5) {0/200};
\end{tikzpicture}%
\caption{\textbf{Revising ownership repairs most replayed errors.} (a) Outcome at the 30th later observation of 100 replayed wrong ownerships under the same evidence. (b) Repairs among them and breaks among 200 events that start correct.}
\label{fig:repair}
\end{gridfloat}
\end{figure}
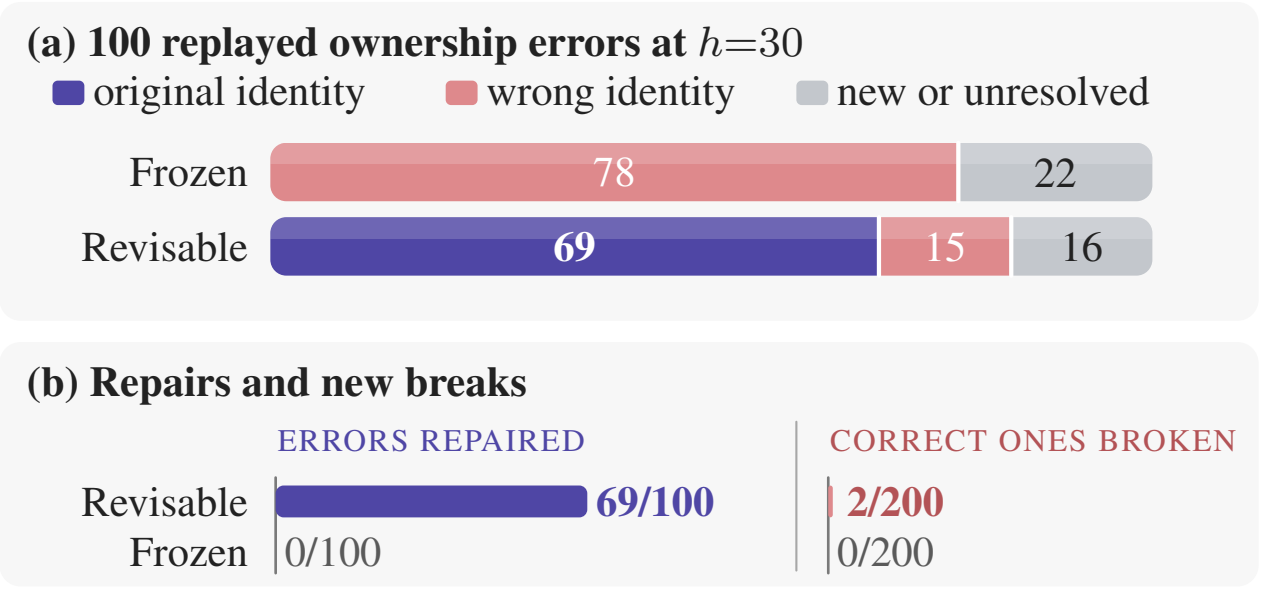
\begin{table*}[t]
\begin{gridfloat}
\centering
\caption{\textbf{CoWeave leads every compared pipeline on all four datasets.} Means over three training seeds, with standard deviations ${}\le{}$\SdMax{} (Supp. Tables S1 and S18). Correct, Wrong, and New/unres.\ partition all \Arrivals{} MTMMC arrivals, and $P_{30}$ uses the \Complete{} with 30 later observations. Time is per online step, including selection, association, and state updates, and relative memory sets Independent${}=1$. $\Delta$ is CoWeave minus GMT (causal) in percentage points, with 95\% interval half-widths in grey, from a paired bootstrap over test episodes on MTMMC and a paired $t$ interval over the three seeds elsewhere. Bold and underline mark the best and second best in each column.}
\label{tab:main}\footnotesize\renewcommand{\arraystretch}{0.87}%
\def\MainBody{\toprule
 & \multicolumn{6}{c}{\textbf{MTMMC}} && \multicolumn{2}{c}{\textbf{Cost}} && \multicolumn{2}{c}{\textbf{WILDTRACK}} && \multicolumn{2}{c}{\textbf{MMPTrack}} && \multicolumn{2}{c}{\textbf{DIVOTrack}}\\
\cmidrule(lr){2-7}\cmidrule(lr){9-10}\cmidrule(lr){12-13}\cmidrule(lr){15-16}\cmidrule(lr){18-19}
 &&&&&&&& Time & Relative &&&&&&&&&\\
\textbf{Method} & G-IDF1$\uparrow$ & HOTA$\uparrow$ & Correct$\uparrow$ & Wrong$\downarrow$ & New/unres.$\downarrow$ & $P_{30}\!\uparrow$ && (ms) & memory && G-IDF1$\uparrow$ & $P_{30}\!\uparrow$ && G-IDF1$\uparrow$ & $P_{30}\!\uparrow$ && G-IDF1$\uparrow$ & $P_{30}\!\uparrow$\\\midrule
\RowsMainwide
\bottomrule}%
\setlength{\tabcolsep}{0pt}\sbox0{\begin{tabular}{@{}l*{6}{c}c*{2}{c}c*{2}{c}c*{2}{c}c*{2}{c}@{}}\MainBody\end{tabular}}%
\setlength{\tabcolsep}{\dimexpr(\linewidth-\wd0)/36\relax}%
\begin{tabular}{@{}l*{6}{c}c*{2}{c}c*{2}{c}c*{2}{c}c*{2}{c}@{}}\MainBody\end{tabular}
\end{gridfloat}
\end{table*}
\subsection{Persistence}
\label{sec:persistence}

Throughout a read loop, ownership and relation validity stay fixed, so the losses before and after a read use one interpretation of the records, and revisions begin only after emission. Once $D_S$ is emitted, each new record receives an ownership distribution from the feasible assignment posterior. For a record $r$ created from observation $i_r$, ownership is the posterior marginal of the identity assigned to $i_r$,
\begin{equation}\gridfix{-0.44pt}
\label{eq:ownership}
\pi_r(j)=\sum_{H\in\mathcal H_t}P_S(H)\,\ind\bigl(H(i_r)=j\bigr),
\end{equation}
where $j$ ranges over the existing identities and $\pi_r(\varnothing)$ takes the mass of fresh labels. As $H$ has no unresolved value, an unresolved output belongs to $D_S$ alone and leaves $\pi_r$ at the posterior marginal specified by Eq.~\ref{eq:ownership}.

The owner of $r$ is one variable, the identity of observation $i_r$, and stays open to revision. When a later step reads the source of $r$, the terms of Eq.~\ref{eq:source_support} that link $z_r$ to the current observations join the factor log of $r$, and after emission $\pi_r$ is recomputed as the marginal of this variable over all logged factors, each counted once, with the current observations summed out under $P_S$. A dependency index marks the summaries built from the record, whose recomputed versions enter the base scores of later arrivals. If an image of B was written to A, a later clear view of B that reads its source thus moves the record to B and corrects both summaries, as Figure~\ref{fig:repair} shows in a paired replay. A validity head trained with $\mathcal L_{\mathrm{valid}}$ in Eq.~\ref{eq:training} predicts $v_{rs}(t)$, so a group split lowers $v_{rs}(t)$ without touching either ownership. Emitted decisions stay immutable, so a revision changes only the evidence that later arrivals read, never an identity that has already been emitted for an earlier arrival.

Because memory is bounded, CoWeave retains records by the increase in predicted future decision risk per byte that deleting them would cause, so a rare distinguishing detail outweighs a recent redundant descriptor, and it removes source bundles with their dependent records, so no summary keeps evidence from a deleted source.

\subsection{Learning}
\label{sec:learning}

Training aligns the predictors with the loss that decisions and reads use. We train scoring, validity, and value on causal episodes whose predicted tracks and historical assignments form the candidates, with ground-truth identities supervising losses after input construction. The objective combines feasible-assignment likelihood, relation validity supervision, and prediction of each source's read value,
\begin{equation}\gridfix{4.71pt}
\label{eq:training}
\begin{split}
\mathcal L={}&-\log P_S(H^*)+\lambda_v\mathcal L_{\mathrm{valid}}\\
&+\lambda_q\lVert\widehat V(u\mid S)-V^*(u\mid S)\rVert_2^2.
\end{split}
\end{equation}
Here $H^*$ is the reference assignment, $\widehat V$ the predicted read value, $\lambda_v$ and $\lambda_q$ weight the terms, and $\mathcal L_{\mathrm{valid}}$ is a binary classification loss on $v_{rs}(t)$ labeled from known valid and invalid relations. The assignment likelihood is supervised only when $H^*$ survives candidate generation and the retained hypothesis search. Otherwise, this likelihood term is masked, and the case counts as a coverage failure. The read-value target is $V^*=L(D_S,H^*)-L(D_{S\cup\{u\}},H^*)$, using a legal training source and identical decision costs. Ambiguous relation labels are masked, and development data set loss weights and calibration. The decision loss $L$ of Eq.~\ref{eq:loss} scores emitted decisions, whereas $\mathcal L$ trains the functions behind them: its likelihood term shapes $P_S$ and hence $D_S$ and the ownership of Eq.~\ref{eq:ownership}, its validity term trains $v_{rs}(t)$, and its value term regresses the reduction of $L$, the same loss with which retention estimates the risk of future decisions.

\section{Evaluation}
\label{sec:experiments}
We ask whether CoWeave continues identities correctly at arrivals and keeps them over later observations, which of its modules produce the differences, and how its lead depends on the shared budget and on invalid relations.
\begin{figure*}[t]
\begin{gridfloat}
\centering
\begin{tikzpicture}[cw,font=\footnotesize]
\path (0,0) (496.8,92);
\cwpcard{0}{0}{272}{92}{(a) $P_k$ after arrival}
\cwpcard{276}{0}{496.8}{92}{(b) HOTA by reader}
\node[anchor=north west,inner sep=0pt,font=\footnotesize\bfseries,text=cwInk] at (393,87.5) {(c) $P_{30}$ by reader};
\begin{axis}[cw axis, at={(30pt,24pt)}, width=190pt, height=48pt,
  xmin=-1, xmax=31, ymin=55, ymax=90, xtick={0,5,10,20,30}, ytick={60,70,80,90},
  xlabel={later observations $k$}, ylabel={$P_k$ (\%)}]
  \addplot[name path=ours, draw=none, forget plot] coordinates {(0,86.5) (5,85.5) (10,84.6) (20,83.1) (30,82.1)};
  \addplot[name path=gmt,  draw=none, forget plot] coordinates {(0,77.8) (5,75.4) (10,73.2) (20,69.6) (30,67.0)};
  \addplot[cwOursL, forget plot] fill between[of=ours and gmt];
  \addplot[cwL4, line width=0.8pt, mark=*, mark size=1.2pt, mark options={fill=cwL4,draw=white,line width=0.3pt}] coordinates {(0,70.3) (5,67.8) (10,65.3) (20,61.4) (30,58.6)};
  \addplot[cwL3, line width=0.8pt, mark=diamond*, mark size=1.6pt, mark options={fill=cwL3,draw=white,line width=0.3pt}] coordinates {(0,72.7) (5,70.4) (10,68.2) (20,64.7) (30,62.2)};
  \addplot[cwB2, line width=0.8pt, mark=triangle*, mark size=1.6pt, mark options={fill=cwB2,draw=white,line width=0.3pt}] coordinates {(0,74.5) (5,72.2) (10,70.1) (20,66.7) (30,64.2)};
  \addplot[cwB1, line width=1.1pt, mark=square*, mark size=1.3pt, mark options={fill=cwB1,draw=white,line width=0.3pt}] coordinates {(0,77.8) (5,75.4) (10,73.2) (20,69.6) (30,67.0)};
  \addplot[cwOurs, line width=4.2pt, opacity=0.16, forget plot] coordinates {(0,86.5) (5,85.5) (10,84.6) (20,83.1) (30,82.1)};
  \addplot[cwOurs, line width=1.6pt, mark=*, mark size=1.8pt, mark options={fill=cwOurs,draw=white,line width=0.5pt}] coordinates {(0,86.5) (5,85.5) (10,84.6) (20,83.1) (30,82.1)};
\end{axis}
\pgfmathsetmacro{\yu}{48/35}
\def\cwoursname{CoWeave}
\foreach \v/\l/\lab/\cl/\f in {82.1/82.1/CoWeave/cwOurs/\bfseries,67.0/71.5/GMT/cwB1/{},64.2/65.0/MCTR/cwB2/{},62.2/58.4/ReST/cwB3!75!black/{},58.6/51.8/DyGLIP/cwB4!60!black/{}}{
  \edef\cwlab{\lab}\ifx\cwlab\cwoursname\else\draw[\cl,line width=0.4pt] (30+190+2.5,{24+(\v-55)*\yu}) -- (30+190+8,{24+(\l-55)*\yu});\fi
  \node[anchor=west,inner sep=1pt,text=\cl,font=\footnotesize\f] at (30+190+8,{24+(\l-55)*\yu}) {\lab};}
\foreach \px/\ymin/\ymax/\yt/\ga/\gb/\ca/\cb in {306/59/71/{60,65,70}/61.4/64.6/65.0/69.3, 418/63/86/{64,72,80}/66.3/73.4/73.6/82.1}{
\begin{axis}[cw axis, at={(\px pt,24pt)}, width=62pt, height=38pt,
  xmin=-0.4, xmax=1.4, ymin=\ymin, ymax=\ymax, xtick={0,1}, xticklabels={tracklet,CoWeave}, ytick=\yt,
  xlabel={reader}]
  \addplot[name path=m1, draw=none, forget plot] coordinates {(0,\ca) (1,\cb)};
  \addplot[name path=m0, draw=none, forget plot] coordinates {(0,\ga) (1,\gb)};
  \addplot[cwOursL, forget plot] fill between[of=m1 and m0];
  \addplot[cwB1, line width=1.1pt, mark=square*, mark size=1.6pt, mark options={fill=cwB1,draw=white,line width=0.3pt}] coordinates {(0,\ga) (1,\gb)};
  \addplot[cwOurs, line width=4.2pt, opacity=0.16, forget plot] coordinates {(0,\ca) (1,\cb)};
  \addplot[cwOurs, line width=1.6pt, mark=*, mark size=2pt, mark options={fill=cwOurs,draw=white,line width=0.5pt}] coordinates {(0,\ca) (1,\cb)};
  \node[anchor=north,inner sep=1pt,yshift=-2pt,text=cwB1,font=\fontsize{6}{7}\selectfont] at (axis cs:0,\ga) {\ga};
  \node[anchor=north,inner sep=1pt,yshift=-2pt,text=cwB1,font=\fontsize{6}{7}\selectfont] at (axis cs:1,\gb) {\gb};
  \node[anchor=south,inner sep=1pt,yshift=2pt,text=cwOurs,font=\fontsize{6}{7}\selectfont\bfseries] at (axis cs:0,\ca) {\ca};
  \node[anchor=south,inner sep=1pt,yshift=2pt,text=cwOurs,font=\fontsize{6}{7}\selectfont\bfseries] at (axis cs:1,\cb) {\cb};
\end{axis}}
\node[draw=cwInk!30,fill=white,rounded corners=2pt,line width=0.4pt,inner xsep=3.5pt,inner ysep=1.6pt] at (386.4,74) {%
  \tikz[baseline=-0.6ex]{\draw[cwOurs,line width=1.6pt] (0pt,0pt) -- (10pt,0pt);
    \fill[cwOurs,draw=white,line width=0.4pt] (5pt,0pt) circle[radius=2pt];}\,\textcolor{cwOurs}{CoWeave matcher}\hspace{9pt}%
  \tikz[baseline=-0.6ex]{\draw[cwB1,line width=1.1pt] (0pt,0pt) -- (10pt,0pt);
    \path[fill=cwB1,draw=white,line width=0.3pt] (3.4pt,-1.6pt) rectangle ++(3.2pt,3.2pt);}\,\textcolor{cwB1}{GMT adapter}};
\end{tikzpicture}%
\caption{\textbf{CoWeave's lead holds over later observations, and the reader and the matcher each raise it.} (a) Share $P_k$ of arrivals with $k$ later observations correct throughout, computed over the arrivals that have $k$ later observations; the band is the lead over GMT. (b, c) HOTA and $P_{30}$ of two source readers crossed with two matchers under a common front end.}
\label{fig:results}
\end{gridfloat}
\end{figure*}
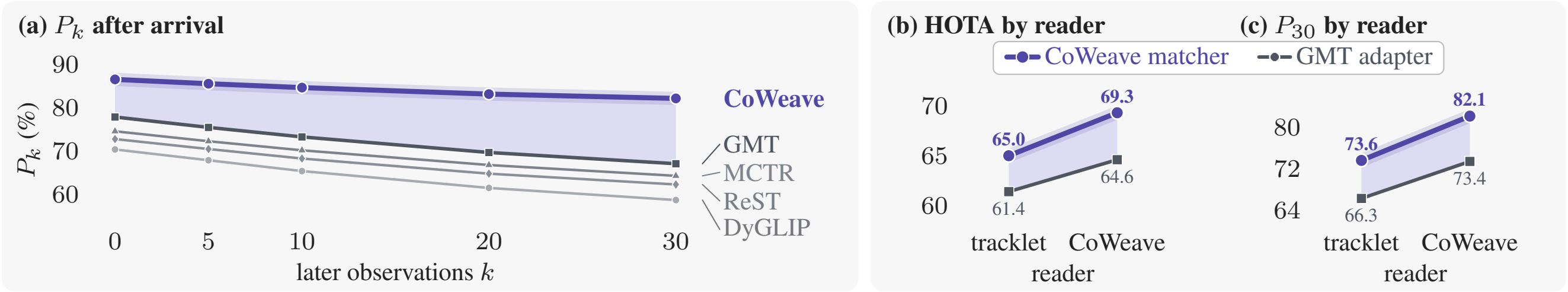
\subsection{Setup}
\label{sec:protocol}
\runin{Dataset.}We evaluate on MTMMC, which provides synchronized RGB streams and a global person identity across cameras~\cite{woo2024mtmmc}. Training, development, and test sequences do not overlap. The \Episodes{} test episodes contain \Arrivals{} eligible arrivals, of which \Complete{} have 30 later observations. All complete-system comparisons share a camera schedule, an output deadline, a published observation prefix, and a budget of \BudgetCap{} units per decision step for reading sources and scoring candidates, shared by all active people.

\runin{Baselines.}We compare CoWeave with DyGLIP~\cite{quach2021dyglip} and ReST~\cite{cheng2023rest}, which associate tracklets and detections, respectively, through relation graphs, with MCTR~\cite{niculescumizil2025mctr}, which maintains probabilistic view-to-global associations, and with GMT~\cite{zhen2026gmt}, which combines cross-view histories, neighboring-target relations, and occlusion memory. In Table~\ref{tab:main}, all methods share detections, descriptors, candidates, source pool, and read budget, and GMT (causal) runs GMT on the observations available at decision time. Independent reads stored history under the same budget but scores each candidate identity alone, with no relations between people and no revision of the ownership of stored observations.

\runin{Metrics.}Because sequence-level scores average over whole sequences, we take arrival outcomes and uninterrupted persistence as the primary measures. An arrival occurs when a person enters a camera view after being observed in another camera. At each arrival, a frozen one-to-one map from previously emitted identities to reference identities, built from earlier published boxes at an intersection over union (IoU) of at least 0.5, decides whether the arrival continues the person's established identity, which counts as correct, or gives an established identity to another person, which counts as wrong. New labels, unresolved outputs, missed detections, and missing anchors form the third class, New/unres., so the three shares partition all eligible arrivals. Persistence $P_k$ is the share of arrivals with $k$ later observations that are correct at arrival and at each of these $k$ observations. Thus $P_0$ equals the correct share, and $k$ counts observations of the person, not frames or seconds. G-IDF1 applies one global tracker-to-reference identity mapping following~\cite{ristani2016performance} and equals $2\,\mathrm{IDTP}/(2\,\mathrm{IDTP}+\mathrm{IDFP}+\mathrm{IDFN})$ over identity true positives, false positives, and false negatives, and HOTA scores detection and association jointly~\cite{luiten2021hota}.

\subsection{Identity continuity}
CoWeave improves identity continuation while reducing wrong arrivals. Table~\ref{tab:main} shows reading history without relations gives \IndepW\% wrong arrivals, adding context and cross-view memory lowers this to \BaseW\% for GMT, and CoWeave lowers it further to \GoalW\%, a drop of \Wdelta{} points or about \Wrelative\% relative to GMT, while its correct share rises from \BaseC\% to \GoalC\%. Because the remaining arrivals stay in the same denominator and their share also falls, from \BaseN\% to \GoalN\%, CoWeave commits to more arrivals while making fewer errors. All 95\% intervals marked in Table~\ref{tab:main} exclude zero, on MTMMC and on the three other datasets alike, as detailed separately for each dataset in Supp. Table S14.
\begin{table}[t]
\begin{gridfloat}
\centering
\caption{\textbf{The lead holds with each method's own front end.} MTMMC results, with per-step cost relative to Independent${}=1$ and $\Delta$ as CoWeave minus GMT (causal) in percentage points.}
\label{tab:budget}\footnotesize\renewcommand{\arraystretch}{0.92}
\setlength{\tabcolsep}{4pt}
\begin{tabularx}{\linewidth}{l*{3}{>{\centering\arraybackslash}X}}\toprule
\textbf{Method} & G-IDF1$\uparrow$ & $P_{30}\uparrow$ & Relative cost\\\midrule
\RowsFrontend
\bottomrule\end{tabularx}
\end{gridfloat}
\end{table}
\begin{table}[t]
\begin{gridfloat}
\centering
\caption{\textbf{Every part and the retention rule matter on MTMMC.} Part replacements share one read budget and retention policies one storage cap. $\Delta P_{30}$ is the change in percentage points relative to the full configuration within the corresponding comparison group.}
\label{tab:ablation-main}\footnotesize\renewcommand{\arraystretch}{0.92}
\setlength{\tabcolsep}{3pt}
\begin{tabularx}{\linewidth}{Xcccc}\toprule
\textbf{Configuration} & G-IDF1$\uparrow$ & Wrong$\downarrow$ & $P_{30}\uparrow$ & $\Delta P_{30}$\\\midrule
\RowsAblationmain
\bottomrule\end{tabularx}
\end{gridfloat}
\end{table}

The lead persists with each pipeline's own front end. As Table~\ref{tab:budget} shows, when every pipeline runs its own front end, CoWeave reaches \NativeCo{} G-IDF1 against \NativeGmt{} for GMT, a lead of \NativeDelta{} G-IDF1 and \NativeDeltaP{} $P_{30}$ points at a step cost below that of MCTR. The five baselines also keep the order they have under the shared front end. Under the shared front end, its \GoalID{} G-IDF1 also exceeds OCMCTrack~\cite{specker2024ocmctrack} with additional geometry (76.0) and lifted multicuts~\cite{nguyen2022lmgp} with future tracklets (78.7), both of which use more information.

CoWeave improves both arrival accuracy and uninterrupted correctness over the later observations. It keeps \GoalP\% of the arrivals with 30 later observations correct throughout, against \BaseP\% for GMT, a gain of \DeltaP{} points, and Figure~\ref{fig:results}(a) traces the gap over $k$. CoWeave reaches this at a mean step time of \GoalMs{} ms, \TimeDelta\% above GMT and below the \MctrMs{} ms of MCTR, while retaining \StoreSave\% less stored state than the Independent baseline.

The lead carries over to other multi-view datasets. Table~\ref{tab:main} applies the protocol to WILDTRACK~\cite{chavdarova2018wildtrack}, MMPTrack~\cite{han2023mmptrack}, and DIVOTrack~\cite{hao2024divotrack}, which add shared identities at different scales, indoor occlusion, and viewpoint and scene variation. On each, CoWeave is the best of all six pipelines, and its $P_{30}$ gain of \TransferPMin{} to \TransferPMax{} points over GMT is about twice its G-IDF1 gain, as on MTMMC. On these datasets, its wrong-arrival fraction is \TransWcMin\% to \TransWcMax\%, about half of the \TransWgMin\% to \TransWgMax\% that GMT produces, and its HOTA is \TransHMin{} to \TransHMax{} points higher.

\begin{figure*}[t]
\begin{gridfloat}[1]
\centering
\begin{tikzpicture}[cw,font=\footnotesize]
\path (0,0) (496.8,92);
\cwpcard{0}{0}{214}{92}{(a) Correct arrivals by evidence condition}
\cwpcard{218}{0}{356}{92}{(b) Shared budget}
\cwpcard{360}{0}{496.8}{92}{(c) Invalid relations}
\fill[cwB2,draw=white,line width=0.4pt] (36,76) circle[radius=2.2pt]; \node[anchor=west,text=cwB1,inner sep=1pt] at (39.5,76) {GMT (causal)};
\fill[cwOurs,draw=white,line width=0.4pt] (100,76) circle[radius=2.4pt]; \node[anchor=west,text=cwOurs,inner sep=1pt] at (103.5,76) {CoWeave};
\draw[cwOursL,line width=3.4pt,line cap=round] (148,74) -- (148,78.5); \node[anchor=west,text=cwOurs,inner sep=1pt] at (152,76) {lead};
\begin{axis}[cw axis, at={(34pt,26pt)}, width=168pt, height=42pt,
  ymin=40, ymax=100, ytick={40,60,80,100},
  xtick=data, symbolic x coords={reliable,shared group,after split,invalid},
  xticklabels={reliable,{shared\\[-2pt]group},{after\\[-2pt]split},invalid},
  enlarge x limits=0.14, ylabel={Correct (\%)},
  x tick label style={font=\footnotesize,text=cwInk,yshift=1pt,align=center}]
  \addplot[draw=none, forget plot] coordinates {(reliable,89.7) (shared group,65.4) (after split,59.5) (invalid,51.6)};
  \foreach \c/\g/\o in {reliable/89.7/95.6,shared group/65.4/84.8,after split/59.5/84.1,invalid/51.6/83.2}{
    \edef\cwtmp{\noexpand\draw[cwOursL,line width=5.2pt,line cap=round] (axis cs:\c,\g) -- (axis cs:\c,\o);
      \noexpand\fill[cwB2,draw=white,line width=0.5pt] (axis cs:\c,\g) circle[radius=2.6pt];
      \noexpand\fill[cwOurs,draw=white,line width=0.5pt] (axis cs:\c,\o) circle[radius=2.9pt];}\cwtmp}
\end{axis}
\begin{axis}[cw axis, at={(246pt,26pt)}, width=66pt, height=46pt,
  xmode=log, log basis x=2, xmin=0.2, xmax=2.5, xtick={0.25,0.5,1,2}, xticklabels={0.25,0.5,1,2},
  ymin=60, ymax=85, ytick={60,70,80}, xlabel={budget ($\times$ default)}, ylabel={G-IDF1 (\%)}]
  \addplot[name path=ours, draw=none, forget plot] coordinates {(0.25,74.2) (0.5,77.4) (1,80.5) (2,81.4)};
  \addplot[name path=gmt, draw=none, forget plot] coordinates {(0.25,65.1) (0.5,69.3) (1,72.9) (2,74.2)};
  \addplot[cwOursL, forget plot] fill between[of=ours and gmt];
  \addplot[cwL4, line width=0.8pt, mark=*, mark size=1.2pt, mark options={fill=cwL4,draw=white,line width=0.3pt}] coordinates {(0.25,63.1) (0.5,66.5) (1,68.8) (2,70.0)};
  \addplot[cwB1, line width=1.1pt, mark=*, mark size=1.4pt, mark options={fill=cwB1,draw=white,line width=0.3pt}] coordinates {(0.25,65.1) (0.5,69.3) (1,72.9) (2,74.2)};
  \addplot[cwOurs, line width=4.2pt, opacity=0.16, forget plot] coordinates {(0.25,74.2) (0.5,77.4) (1,80.5) (2,81.4)};
  \addplot[cwOurs, line width=1.6pt, mark=*, mark size=1.8pt, mark options={fill=cwOurs,draw=white,line width=0.5pt}] coordinates {(0.25,74.2) (0.5,77.4) (1,80.5) (2,81.4)};
\end{axis}
\node[anchor=west,text=cwOurs,font=\footnotesize\bfseries,inner sep=1pt] at (315,{26+(81.4-60)*46/25+1}) {CoWeave};
\node[anchor=west,text=cwB1,inner sep=1pt] at (315,{26+(74.2-60)*46/25+1}) {GMT};
\node[anchor=west,text=cwB3!80!black,inner sep=1pt] at (315,{26+(70.0-60)*46/25-2.5}) {Indep.};
\begin{axis}[cw axis, at={(388pt,26pt)}, width=62pt, height=46pt,
  xmin=-3, xmax=43, xtick={0,10,20,30,40}, ymin=60, ymax=85, ytick={60,70,80},
  xlabel={invalid relations (\%)}, ylabel={G-IDF1 (\%)}]
  \addplot[name path=ours, draw=none, forget plot] coordinates {(0,80.5) (10,80.0) (20,79.4) (30,78.6) (40,77.6)};
  \addplot[name path=gmt, draw=none, forget plot] coordinates {(0,72.9) (10,71.4) (20,69.4) (30,67.6) (40,65.8)};
  \addplot[cwOursL, forget plot] fill between[of=ours and gmt];
  \addplot[cwL4, line width=0.8pt, mark=*, mark size=1.2pt, mark options={fill=cwL4,draw=white,line width=0.3pt}] coordinates {(0,68.8) (10,68.8) (20,68.8) (30,68.8) (40,68.8)};
  \addplot[cwB1, line width=1.1pt, mark=*, mark size=1.4pt, mark options={fill=cwB1,draw=white,line width=0.3pt}] coordinates {(0,72.9) (10,71.4) (20,69.4) (30,67.6) (40,65.8)};
  \addplot[cwOurs, line width=4.2pt, opacity=0.16, forget plot] coordinates {(0,80.5) (10,80.0) (20,79.4) (30,78.6) (40,77.6)};
  \addplot[cwOurs, line width=1.6pt, mark=*, mark size=1.8pt, mark options={fill=cwOurs,draw=white,line width=0.5pt}] coordinates {(0,80.5) (10,80.0) (20,79.4) (30,78.6) (40,77.6)};
\end{axis}
\node[anchor=west,text=cwOurs,font=\footnotesize\bfseries,inner sep=1pt] at (453,{26+(77.6-60)*46/25+1}) {CoWeave};
\node[anchor=west,text=cwB1,inner sep=1pt] at (453,{26+(65.8-60)*46/25-2}) {GMT};
\node[anchor=west,text=cwB3!80!black,inner sep=1pt] at (453,{26+(68.8-60)*46/25+3}) {Indep.};
\end{tikzpicture}%
\caption{\textbf{CoWeave's lead grows where context misleads GMT and holds under a small budget.} (a) Correct share in four evidence conditions: relations reliable and instance detail clear, two people in one group with weak instance detail, reappearance after a group separates, and relations made invalid while instance evidence stays valid. (b) G-IDF1 as the shared budget for reading and selection varies on a log scale. (c) G-IDF1 as an increasing fraction of historical relations is made invalid.}
\label{fig:analysis}
\end{gridfloat}
\end{figure*}
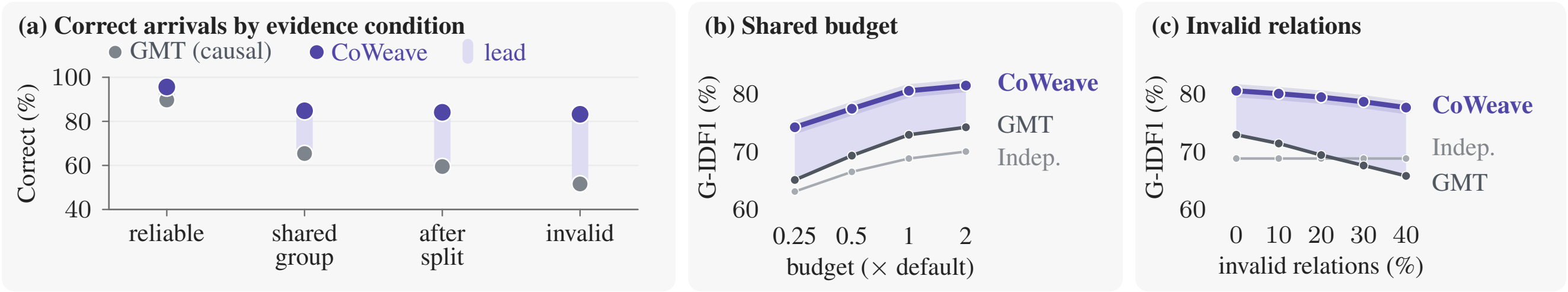
\subsection{Isolating the modules}
Table~\ref{tab:ablation-main} and Figures~\ref{fig:repair} and~\ref{fig:results}(b, c) test, with matched controls, the modules aimed at the two failure modes: the reader and matcher for the score at an arrival, and revisable ownership for the memory it leaves. Under a common front end, Figure~\ref{fig:results}(b, c) crosses two source readers with two association matchers. The tracklet reader ranks admitted sources by similarity to the current observation, whereas the CoWeave reader scores their expected reduction of the decision loss after the sources already selected. Starting from the tracklet reader with the GMT adapter, $P_{30}$ rises by \ReaderGain{} points when only the reader changes, by \MatcherGain{} when only the matcher changes, and by \BothGain{} when both change. Reading by value does not slow the step (\ReaderMs{} against \TrackletMs{} ms with the GMT adapter). The 95\% interval of the interaction, [\InteractionPlo, \InteractionPhi] points, contains zero, as does that of the G-IDF1 interaction, so no clear interaction between them is detected for either of the two evaluation metrics.

Revisable ownership repairs wrong matches that memory would otherwise keep. In a paired replay (Supp. Sec. S5), two branches start from the same erroneous ownership and receive the same later evidence, one with frozen and one with revisable ownership, and a repair counts only when the original identity returns. As Figure~\ref{fig:repair} shows, of 100 replayed errors, revisable ownership returns to the original identity in \RepairEver{}, a median of \RepairMedian{} observations into the replay, and \RepairHeld{} of them still hold it at the 30th later observation. Frozen ownership never returns, and among 200 continuations that start correct, revision breaks 2. Revising single records also beats revising whole tracklets: a control that reassigns whole tracklets under the same reader, budget, and online output rule reaches \RevGranoP\% $P_{30}$ in Table~\ref{tab:ablation-main}, \RevGran{} points below CoWeave, with \RevGranoW\% against \GoalW\% wrong arrivals.

Replacing single parts of the full system at the same budget confirms each of them. As Table~\ref{tab:ablation-main} shows, every replacement lowers $P_{30}$, by up to \AblInstance{} points without instance detail, and lowers G-IDF1 by \AblFixedRetID{} to \AblInstanceID{} points. Permuting relations while keeping each record's number of relations costs \PermCost{} $P_{30}$ points, the same \ZeroCost{} points lost by zeroing relational support, so the relational gain depends on which people a relation links, not on how many links exist. Under the same storage cap, retention by predicted decision risk beats the strongest alternative for each metric by \RetMarginID{} G-IDF1 and \RetMarginP{} $P_{30}$ points. Across one-factor sweeps of $\lambda_U$, $\lambda_B$, $\lambda_{\mathrm{rel}}$, $T$, and the candidate and beam limits, G-IDF1 and $P_{30}$ both stay at or above \SweepMinP{}, above every compared pipeline, as Supp. Table S17 lists.

\subsection{Budget and invalid relations}

The lead is smallest when the evidence is cleanest and widest where context misleads GMT. As Figure~\ref{fig:analysis}(a) shows, CoWeave's correct share exceeds that of GMT by \LeadReliable{} points when relations are reliable and instance detail is clear, by \LeadGroup{} points when people share a group and instance detail is weak, and by \LeadSplit{} and \LeadInvalid{} points after a group separates and when relations are invalid. When nothing distinguishes the competing people, it abstains: in all \SymArrivals{} arrivals of \SymEpisodes{} episodes with complete symmetry, it leaves the identity unresolved and gives no established identity to the wrong person, as Supp. Table S8 lists.

The lead holds when the budget shrinks and relations decay. In Figure~\ref{fig:analysis}(b), CoWeave reaches \QuarterCo{} G-IDF1 at a quarter of the default budget, above what GMT reaches with the full budget and level with GMT at twice the budget (\DoubleGmt{}). At that quarter budget, CoWeave still keeps \QuarterCoP\% of the arrivals with 30 later observations correct throughout, above the \BaseP\% of GMT with the full budget in Table~\ref{tab:main}. When 40\% of the relations are made invalid, CoWeave loses \InvLossCo{} G-IDF1 points and GMT loses \InvLossGmt{}, as Figure~\ref{fig:analysis}(c) shows, which is consistent with relation validity discounting historical relations that no longer hold at the current arrival.

\section{Conclusion}
\begingroup\parfillskip=0pt
\label{sec:conclusion}
CoWeave unifies assignment scoring, retrieval, and memory through one identity decision loss. It compares competing assignments with candidate-specific evidence, reads history by predicted loss reduction, and keeps each observation's ownership revisable. On MTMMC and three other datasets, under shared and native front ends, it reduces wrong arrivals and improves identity persistence over causal GMT, particularly with shared or stale context. Under the same budget cap, its step time is \TimeDelta\% higher than GMT's, with \StoreSave\% less stored state than Independent. Each stored source can serve several people, while its access cost is charged only once. \emph{Historical context helps when it distinguishes competing identities and its attribution remains revisable as later evidence arrives.}\label{page:main-end}\par\endgroup

\section*{LLM Usage Statement}
\label{page:llm-start}
The authors conceived the research question, designed the method and every experiment and analysis, made all research decisions, and checked every number against the result files. LLM-based assistants (Claude, Codex, Gemini and ChatGPT) were used only as tools to polish the writing. The authors take full responsibility for the content.

{\small\bibliographystyle{ieeenat_fullname}\bibliography{references}}

\begin{thebibliography}{53}
\providecommand{\natexlab}[1]{#1}
\providecommand{\url}[1]{\texttt{#1}}
\expandafter\ifx\csname urlstyle\endcsname\relax
  \providecommand{\doi}[1]{doi: #1}\else
  \providecommand{\doi}{doi: \begingroup \urlstyle{rm}\Url}\fi

\bibitem[Bergmann et~al.(2019)Bergmann, Meinhardt, and
  Leal-Taix{\'e}]{bergmann2019tracktor}
Philipp Bergmann, Tim Meinhardt, and Laura Leal-Taix{\'e}.
\newblock {Tracking Without Bells and Whistles}.
\newblock In \emph{{Proceedings of the IEEE/CVF International Conference on
  Computer Vision (ICCV)}}, pages 941--951, 2019.

\bibitem[Bernardin and Stiefelhagen(2008)]{bernardin2008clear}
Keni Bernardin and Rainer Stiefelhagen.
\newblock {Evaluating Multiple Object Tracking Performance: The {CLEAR MOT}
  Metrics}.
\newblock \emph{EURASIP Journal on Image and Video Processing}, 2008:\penalty0
  246309, 2008.

\bibitem[Bewley et~al.(2016)Bewley, Ge, Ott, Ramos, and
  Upcroft]{bewley2016sort}
Alex Bewley, Zongyuan Ge, Lionel Ott, Fabio Ramos, and Ben Upcroft.
\newblock Simple online and realtime tracking.
\newblock In \emph{IEEE International Conference on Image Processing (ICIP)},
  pages 3464--3468, 2016.

\bibitem[Bras{\'o} and Leal-Taix{\'e}(2020)]{braso2020neural}
Guillem Bras{\'o} and Laura Leal-Taix{\'e}.
\newblock {Learning a Neural Solver for Multiple Object Tracking}.
\newblock In \emph{{Proceedings of the IEEE/CVF Conference on Computer Vision
  and Pattern Recognition (CVPR)}}, pages 6246--6256, 2020.

\bibitem[Cai et~al.(2022)Cai, Xu, Li, Xiong, Xia, Tu, and Soatto]{cai2022memot}
Jiarui Cai, Mingze Xu, Wei Li, Yuanjun Xiong, Wei Xia, Zhuowen Tu, and Stefano
  Soatto.
\newblock {MeMOT: Multi-Object Tracking With Memory}.
\newblock In \emph{{Proceedings of the IEEE/CVF Conference on Computer Vision
  and Pattern Recognition (CVPR)}}, pages 8080--8090, 2022.

\bibitem[Cao et~al.(2023)Cao, Pang, Weng, Khirodkar, and Kitani]{cao2023ocsort}
Jinkun Cao, Jiangmiao Pang, Xinshuo Weng, Rawal Khirodkar, and Kris Kitani.
\newblock {Observation-Centric SORT: Rethinking SORT for Robust Multi-Object
  Tracking}.
\newblock In \emph{{Proceedings of the IEEE/CVF Conference on Computer Vision
  and Pattern Recognition (CVPR)}}, pages 9686--9696, 2023.

\bibitem[Chavdarova et~al.(2018)Chavdarova, Baqu{\'e}, Bouquet, Maksai, Jose,
  Bagautdinov, Lettry, Fua, Van~Gool, and Fleuret]{chavdarova2018wildtrack}
Tatjana Chavdarova, Pierre Baqu{\'e}, St{\'e}phane Bouquet, Andrii Maksai, Cijo
  Jose, Timur Bagautdinov, Louis Lettry, Pascal Fua, Luc Van~Gool, and Fran{\c
  c}ois Fleuret.
\newblock {WILDTRACK}: A multi-camera {HD} dataset for dense unscripted
  pedestrian detection.
\newblock In \emph{IEEE Conference on Computer Vision and Pattern Recognition
  (CVPR)}, pages 5030--5039, 2018.

\bibitem[Cheng et~al.(2023)Cheng, Qiu, Chiang, and Lai]{cheng2023rest}
Cheng-Che Cheng, Min-Xuan Qiu, Chen-Kuo Chiang, and Shang-Hong Lai.
\newblock {ReST}: A reconfigurable spatial-temporal graph model for
  multi-camera multi-object tracking.
\newblock In \emph{IEEE/CVF International Conference on Computer Vision
  (ICCV)}, pages 10051--10060, 2023.

\bibitem[Cheng and Schwing(2022)]{cheng2022xmem}
Ho~Kei Cheng and Alexander~G. Schwing.
\newblock {XMem: Long-Term Video Object Segmentation with an Atkinson-Shiffrin
  Memory Model}.
\newblock In \emph{{European Conference on Computer Vision (ECCV)}}, pages
  640--658, 2022.

\bibitem[Cheng et~al.(2021)Cheng, Tai, and Tang]{cheng2021stcn}
Ho~Kei Cheng, Yu-Wing Tai, and Chi-Keung Tang.
\newblock {Rethinking Space-Time Networks with Improved Memory Coverage for
  Efficient Video Object Segmentation}.
\newblock In \emph{{Advances in Neural Information Processing Systems}}, pages
  11781--11794, 2021.

\bibitem[Chunduri et~al.(2025)Chunduri, Lu, and Arulraj]{chunduri2025tracer}
Pramod Chunduri, Yao Lu, and Joy Arulraj.
\newblock {TRACER}: Efficient object re-identification in networked cameras
  through adaptive query processing, 2025.
\newblock arXiv:2507.09448.

\bibitem[Dendorfer et~al.(2021)Dendorfer, O{\v s}ep, Milan, Schindler, Cremers,
  Reid, Roth, and Leal-Taix{\'e}]{dendorfer2021motchallenge}
Patrick Dendorfer, Aljo{\v s}a O{\v s}ep, Anton Milan, Konrad Schindler, Daniel
  Cremers, Ian Reid, Stefan Roth, and Laura Leal-Taix{\'e}.
\newblock {MOTChallenge: A Benchmark for Single-Camera Multiple Target
  Tracking}.
\newblock \emph{International Journal of Computer Vision}, 129\penalty0
  (4):\penalty0 845--881, 2021.

\bibitem[Eom et~al.(2021)Eom, Lee, Lee, and Ham]{eom2021memory}
Chanho Eom, Geon Lee, Junghyup Lee, and Bumsub Ham.
\newblock Video-based person re-identification with spatial and temporal memory
  networks.
\newblock In \emph{IEEE/CVF International Conference on Computer Vision
  (ICCV)}, pages 12016--12025, 2021.

\bibitem[Fan et~al.(2025)Fan, Qiao, Zhen, Zhao, Fan, and Wang]{fan2025allday}
Huijie Fan, Yu Qiao, Yihao Zhen, Tinghui Zhao, Baojie Fan, and Qiang Wang.
\newblock All-day multi-camera multi-target tracking.
\newblock In \emph{IEEE/CVF Conference on Computer Vision and Pattern
  Recognition (CVPR)}, pages 16892--16901, 2025.

\bibitem[Fortmann et~al.(1983)Fortmann, Bar-Shalom, and
  Scheffe]{fortmann1983jpda}
Thomas~E. Fortmann, Yaakov Bar-Shalom, and Molly Scheffe.
\newblock Sonar tracking of multiple targets using joint probabilistic data
  association.
\newblock \emph{IEEE Journal of Oceanic Engineering}, 8\penalty0 (3):\penalty0
  173--184, 1983.

\bibitem[Gan et~al.(2021)Gan, Han, Yin, Feng, and Wang]{gan2021mvmhat}
Yiyang Gan, Ruize Han, Liqiang Yin, Wei Feng, and Song Wang.
\newblock {Self-supervised Multi-view Multi-Human Association and Tracking}.
\newblock In \emph{{Proceedings of the 29th ACM International Conference on
  Multimedia}}, pages 282--290, 2021.

\bibitem[Gao and Wang(2023)]{gao2023memotr}
Ruopeng Gao and Limin Wang.
\newblock {MeMOTR: Long-Term Memory-Augmented Transformer for Multi-Object
  Tracking}.
\newblock In \emph{{Proceedings of the IEEE/CVF International Conference on
  Computer Vision (ICCV)}}, pages 9867--9876, 2023.

\bibitem[Gao et~al.(2020)Gao, Wang, Lu, and Liu]{gao2020pose}
Shang Gao, Jingya Wang, Huchuan Lu, and Zimo Liu.
\newblock {Pose-Guided Visible Part Matching for Occluded Person ReID}.
\newblock In \emph{{Proceedings of the IEEE/CVF Conference on Computer Vision
  and Pattern Recognition (CVPR)}}, pages 11741--11749, 2020.

\bibitem[Geifman and El-Yaniv(2019)]{geifman2019selectivenet}
Yonatan Geifman and Ran El-Yaniv.
\newblock {{SelectiveNet}: A Deep Neural Network with an Integrated Reject
  Option}.
\newblock In \emph{{Proceedings of the 36th International Conference on Machine
  Learning}}, pages 2151--2159, 2019.

\bibitem[Guo et~al.(2017)Guo, Pleiss, Sun, and Weinberger]{guo2017calibration}
Chuan Guo, Geoff Pleiss, Yu Sun, and Kilian~Q. Weinberger.
\newblock {On Calibration of Modern Neural Networks}.
\newblock In \emph{{Proceedings of the 34th International Conference on Machine
  Learning}}, pages 1321--1330, 2017.

\bibitem[Han et~al.(2023)Han, You, Wang, Zhang, Chu, Hu, Wang, and
  Liu]{han2023mmptrack}
Xiaotian Han, Quanzeng You, Chunyu Wang, Zhizheng Zhang, Peng Chu, Houdong Hu,
  Jiang Wang, and Zicheng Liu.
\newblock {MMPTRACK}: Large-scale densely annotated multi-camera multiple
  people tracking benchmark.
\newblock In \emph{IEEE/CVF Winter Conference on Applications of Computer
  Vision (WACV)}, pages 4849--4858, 2023.

\bibitem[Hao et~al.(2024)Hao, Liu, Zhan, Jin, Liu, Song, Hwang, and
  Wang]{hao2024divotrack}
Shengyu Hao, Peiyuan Liu, Yibing Zhan, Kaixun Jin, Zuozhu Liu, Mingli Song,
  Jenq-Neng Hwang, and Gaoang Wang.
\newblock {DIVOTrack: A Novel Dataset and Baseline Method for Cross-View
  Multi-Object Tracking in DIVerse Open Scenes}.
\newblock \emph{International Journal of Computer Vision}, 132\penalty0
  (4):\penalty0 1075--1090, 2024.

\bibitem[He et~al.(2021)He, Luo, Wang, Wang, Li, and Jiang]{he2021transreid}
Shuting He, Hao Luo, Pichao Wang, Fan Wang, Hao Li, and Wei Jiang.
\newblock {TransReID: Transformer-Based Object Re-Identification}.
\newblock In \emph{{Proceedings of the IEEE/CVF International Conference on
  Computer Vision (ICCV)}}, pages 14993--15002, 2021.

\bibitem[He et~al.(2020)He, Wei, Hong, Shi, and Gong]{he2020tracta}
Yuhang He, Xing Wei, Xiaopeng Hong, Weiwei Shi, and Yihong Gong.
\newblock Multi-target multi-camera tracking by tracklet-to-target assignment.
\newblock \emph{IEEE Transactions on Image Processing}, 29:\penalty0
  5191--5205, 2020.

\bibitem[Jain et~al.(2020)Jain, Zhang, Zhou, Ananthanarayanan, Jiang, Shu,
  Bahl, and Gonzalez]{jain2020spatula}
Samvit Jain, Xun Zhang, Yuhao Zhou, Ganesh Ananthanarayanan, Junchen Jiang,
  Yuanchao Shu, Paramvir Bahl, and Joseph Gonzalez.
\newblock {Spatula}: Efficient cross-camera video analytics on large camera
  networks.
\newblock In \emph{IEEE/ACM Symposium on Edge Computing (SEC)}, pages 110--124,
  2020.

\bibitem[K{\"o}hl et~al.(2020)K{\"o}hl, Specker, Schumann, and
  Beyerer]{kohl2020mta}
Philipp K{\"o}hl, Andreas Specker, Arne Schumann, and J{\"u}rgen Beyerer.
\newblock The {MTA} dataset for multi-target multi-camera pedestrian tracking
  by weighted distance aggregation.
\newblock In \emph{IEEE/CVF Conference on Computer Vision and Pattern
  Recognition Workshops (CVPRW)}, pages 4489--4498, 2020.

\bibitem[Li et~al.(2021)Li, Weng, Xu, and Kitani]{li2021visio}
Yu-Jhe Li, Xinshuo Weng, Yan Xu, and Kris~M. Kitani.
\newblock Visio-temporal attention for multi-camera multi-target association.
\newblock In \emph{IEEE/CVF International Conference on Computer Vision
  (ICCV)}, pages 9814--9824, 2021.

\bibitem[Luiten et~al.(2021)Luiten, O{\v s}ep, Dendorfer, Torr, Geiger,
  Leal-Taix{\'e}, and Leibe]{luiten2021hota}
Jonathon Luiten, Aljo{\v s}a O{\v s}ep, Patrick Dendorfer, Philip Torr, Andreas
  Geiger, Laura Leal-Taix{\'e}, and Bastian Leibe.
\newblock {HOTA}: A higher order metric for evaluating multi-object tracking.
\newblock \emph{International Journal of Computer Vision}, 129\penalty0
  (2):\penalty0 548--578, 2021.

\bibitem[Luna et~al.(2023)Luna, SanMiguel, Mart{\'i}nez, and
  Carballeira]{luna2023gnncca}
Elena Luna, Juan~C. SanMiguel, Jos{\'e}~M. Mart{\'i}nez, and Pablo Carballeira.
\newblock {Graph Neural Networks for Cross-Camera Data Association}.
\newblock \emph{IEEE Transactions on Circuits and Systems for Video
  Technology}, 33\penalty0 (2):\penalty0 589--601, 2023.

\bibitem[Luo et~al.(2019)Luo, Gu, Liao, Lai, and Jiang]{luo2019bag}
Hao Luo, Youzhi Gu, Xingyu Liao, Shenqi Lai, and Wei Jiang.
\newblock {Bag of Tricks and a Strong Baseline for Deep Person
  Re-Identification}.
\newblock In \emph{{Proceedings of the IEEE/CVF Conference on Computer Vision
  and Pattern Recognition (CVPR) Workshops}}, pages 1487--1495, 2019.

\bibitem[Meinhardt et~al.(2022)Meinhardt, Kirillov, Leal-Taix\'e, and
  Feichtenhofer]{meinhardt2022trackformer}
Tim Meinhardt, Alexander Kirillov, Laura Leal-Taix\'e, and Christoph
  Feichtenhofer.
\newblock {TrackFormer: Multi-Object Tracking With Transformers}.
\newblock In \emph{{Proceedings of the IEEE/CVF Conference on Computer Vision
  and Pattern Recognition (CVPR)}}, pages 8834--8844, 2022.

\bibitem[Nguyen et~al.(2022)Nguyen, Henschel, Rosenhahn, Sonntag, and
  Swoboda]{nguyen2022lmgp}
Duy M.~H. Nguyen, Roberto Henschel, Bodo Rosenhahn, Daniel Sonntag, and Paul
  Swoboda.
\newblock {LMGP}: Lifted multicut meets geometry projections for multi-camera
  multi-object tracking.
\newblock In \emph{IEEE/CVF Conference on Computer Vision and Pattern
  Recognition (CVPR)}, pages 8856--8865, 2022.

\bibitem[Niculescu-Mizil et~al.(2025)Niculescu-Mizil, Patel, and
  Melvin]{niculescumizil2025mctr}
Alexandru Niculescu-Mizil, Deep Patel, and Iain Melvin.
\newblock {MCTR}: Multi camera tracking transformer.
\newblock In \emph{IEEE/CVF Winter Conference on Applications of Computer
  Vision Workshops (WACVW)}, pages 816--826, 2025.

\bibitem[Oh et~al.(2019)Oh, Lee, Xu, and Kim]{oh2019stm}
Seoung~Wug Oh, Joon-Young Lee, Ning Xu, and Seon~Joo Kim.
\newblock {Video Object Segmentation Using Space-Time Memory Networks}.
\newblock In \emph{{Proceedings of the IEEE/CVF International Conference on
  Computer Vision (ICCV)}}, pages 9225--9234, 2019.

\bibitem[Pang et~al.(2021)Pang, Qiu, Li, Chen, Li, Darrell, and
  Yu]{pang2021quasidense}
Jiangmiao Pang, Linlu Qiu, Xia Li, Haofeng Chen, Qi Li, Trevor Darrell, and
  Fisher Yu.
\newblock {Quasi-Dense Similarity Learning for Multiple Object Tracking}.
\newblock In \emph{{Proceedings of the IEEE/CVF Conference on Computer Vision
  and Pattern Recognition (CVPR)}}, pages 164--173, 2021.

\bibitem[Quach et~al.(2021)Quach, Nguyen, Le, Truong, Duong, Tran, and
  Luu]{quach2021dyglip}
Kha~Gia Quach, Pha Nguyen, Huu Le, Thanh-Dat Truong, Chi~Nhan Duong, Minh-Triet
  Tran, and Khoa Luu.
\newblock {DyGLIP}: A dynamic graph model with link prediction for accurate
  multi-camera multiple object tracking.
\newblock In \emph{IEEE/CVF Conference on Computer Vision and Pattern
  Recognition (CVPR)}, pages 13779--13788, 2021.

\bibitem[Reid(1979)]{reid1979mht}
Donald~B. Reid.
\newblock An algorithm for tracking multiple targets.
\newblock \emph{IEEE Transactions on Automatic Control}, 24\penalty0
  (6):\penalty0 843--854, 1979.

\bibitem[Ristani and Tomasi(2018)]{ristani2018features}
Ergys Ristani and Carlo Tomasi.
\newblock Features for multi-target multi-camera tracking and
  re-identification.
\newblock In \emph{IEEE Conference on Computer Vision and Pattern Recognition
  (CVPR)}, pages 6036--6046, 2018.

\bibitem[Ristani et~al.(2016)Ristani, Solera, Zou, Cucchiara, and
  Tomasi]{ristani2016performance}
Ergys Ristani, Francesco Solera, Roger Zou, Rita Cucchiara, and Carlo Tomasi.
\newblock Performance measures and a data set for multi-target, multi-camera
  tracking.
\newblock In \emph{Computer Vision -- ECCV 2016 Workshops}, pages 17--35, 2016.

\bibitem[Seidenschwarz et~al.(2023)Seidenschwarz, Bras\'o, Serrano, Elezi, and
  Leal-Taix\'e]{seidenschwarz2023ghost}
Jenny Seidenschwarz, Guillem Bras\'o, V{\'\i}ctor~Castro Serrano, Ismail Elezi,
  and Laura Leal-Taix\'e.
\newblock {Simple Cues Lead to a Strong Multi-Object Tracker}.
\newblock In \emph{{Proceedings of the IEEE/CVF Conference on Computer Vision
  and Pattern Recognition (CVPR)}}, pages 13813--13823, 2023.

\bibitem[Specker(2024)]{specker2024ocmctrack}
Andreas Specker.
\newblock {OCMCTrack}: Online multi-target multi-camera tracking with
  corrective matching cascade.
\newblock In \emph{IEEE/CVF Conference on Computer Vision and Pattern
  Recognition Workshops (CVPRW)}, pages 7236--7244, 2024.

\bibitem[Sun et~al.(2022)Sun, Cao, Jiang, Yuan, Bai, Kitani, and
  Luo]{sun2022dancetrack}
Peize Sun, Jinkun Cao, Yi Jiang, Zehuan Yuan, Song Bai, Kris Kitani, and Ping
  Luo.
\newblock {DanceTrack: Multi-Object Tracking in Uniform Appearance and Diverse
  Motion}.
\newblock In \emph{{Proceedings of the IEEE/CVF Conference on Computer Vision
  and Pattern Recognition (CVPR)}}, pages 20961--20970, 2022.

\bibitem[Sun et~al.(2018)Sun, Zheng, Yang, Tian, and Wang]{sun2018pcb}
Yifan Sun, Liang Zheng, Yi Yang, Qi Tian, and Shengjin Wang.
\newblock {Beyond Part Models: Person Retrieval with Refined Part Pooling (and
  A Strong Convolutional Baseline)}.
\newblock In \emph{{Proceedings of the European Conference on Computer Vision
  (ECCV)}}, pages 501--518, 2018.

\bibitem[Tang et~al.(2019)Tang, Naphade, Liu, Yang, Birchfield, Wang, Kumar,
  Anastasiu, and Hwang]{tang2019cityflow}
Zheng Tang, Milind Naphade, Ming-Yu Liu, Xiaodong Yang, Stan Birchfield, Shuo
  Wang, Ratnesh Kumar, David Anastasiu, and Jenq-Neng Hwang.
\newblock {CityFlow: A City-Scale Benchmark for Multi-Target Multi-Camera
  Vehicle Tracking and Re-Identification}.
\newblock In \emph{{Proceedings of the IEEE/CVF Conference on Computer Vision
  and Pattern Recognition (CVPR)}}, pages 8789--8798, 2019.

\bibitem[Wojke et~al.(2017)Wojke, Bewley, and Paulus]{wojke2017deep}
Nicolai Wojke, Alex Bewley, and Dietrich Paulus.
\newblock Simple online and realtime tracking with a deep association metric.
\newblock In \emph{IEEE International Conference on Image Processing (ICIP)},
  pages 3645--3649, 2017.

\bibitem[Woo et~al.(2024)Woo, Park, Shin, Kim, and Kweon]{woo2024mtmmc}
Sanghyun Woo, Kwanyong Park, Inkyu Shin, Myungchul Kim, and In~So Kweon.
\newblock {MTMMC}: A large-scale real-world multi-modal camera tracking
  benchmark.
\newblock In \emph{IEEE/CVF Conference on Computer Vision and Pattern
  Recognition (CVPR)}, pages 22335--22346, 2024.

\bibitem[Zeng et~al.(2022)Zeng, Dong, Zhang, Wang, Zhang, and
  Wei]{zeng2022motr}
Fangao Zeng, Bin Dong, Yuang Zhang, Tiancai Wang, Xiangyu Zhang, and Yichen
  Wei.
\newblock {MOTR: End-to-End Multiple-Object Tracking with Transformer}.
\newblock In \emph{{European Conference on Computer Vision (ECCV)}}, pages
  659--675, 2022.

\bibitem[Zhang et~al.(2021)Zhang, Wang, Wang, Zeng, and Liu]{zhang2021fairmot}
Yifu Zhang, Chunyu Wang, Xinggang Wang, Wenjun Zeng, and Wenyu Liu.
\newblock {FairMOT: On the Fairness of Detection and Re-identification in
  Multiple Object Tracking}.
\newblock \emph{International Journal of Computer Vision}, 129\penalty0
  (11):\penalty0 3069--3087, 2021.

\bibitem[Zhang et~al.(2022)Zhang, Sun, Jiang, Yu, Weng, Yuan, Luo, Liu, and
  Wang]{zhang2022bytetrack}
Yifu Zhang, Peize Sun, Yi Jiang, Dongdong Yu, Fucheng Weng, Zehuan Yuan, Ping
  Luo, Wenyu Liu, and Xinggang Wang.
\newblock {ByteTrack}: Multi-object tracking by associating every detection
  box.
\newblock In \emph{European Conference on Computer Vision (ECCV)}, pages 1--21,
  2022.

\bibitem[Zhen et~al.(2026)Zhen, Xu, Wang, Fan, Dong, Zhao, and
  Fan]{zhen2026gmt}
Yihao Zhen, Mingyue Xu, Qiang Wang, Baojie Fan, Jiahua Dong, Tinghui Zhao, and
  Huijie Fan.
\newblock {GMT}: Effective global framework for multi-camera multi-target
  tracking.
\newblock In \emph{IEEE/CVF Conference on Computer Vision and Pattern
  Recognition (CVPR)}, pages 28201--28210, 2026.

\bibitem[Zhou et~al.(2024)Zhou, Pang, and Wang]{zhou2024rmem}
Junbao Zhou, Ziqi Pang, and Yu-Xiong Wang.
\newblock {RMem: Restricted Memory Banks Improve Video Object Segmentation}.
\newblock In \emph{{Proceedings of the IEEE/CVF Conference on Computer Vision
  and Pattern Recognition (CVPR)}}, pages 18602--18611, 2024.

\bibitem[Zhou et~al.(2019)Zhou, Yang, Cavallaro, and Xiang]{zhou2019osnet}
Kaiyang Zhou, Yongxin Yang, Andrea Cavallaro, and Tao Xiang.
\newblock {Omni-Scale Feature Learning for Person Re-Identification}.
\newblock In \emph{{Proceedings of the IEEE/CVF International Conference on
  Computer Vision (ICCV)}}, pages 3701--3711, 2019.

\bibitem[Zhou et~al.(2022)Zhou, Yin, Koltun, and
  Kr\"ahenb\"uhl]{zhou2022global}
Xingyi Zhou, Tianwei Yin, Vladlen Koltun, and Philipp Kr\"ahenb\"uhl.
\newblock {Global Tracking Transformers}.
\newblock In \emph{{Proceedings of the IEEE/CVF Conference on Computer Vision
  and Pattern Recognition (CVPR)}}, pages 8761--8770, 2022.

\end{thebibliography}


\begin{thebibliography}{11}
\providecommand{\natexlab}[1]{#1}
\providecommand{\url}[1]{\texttt{#1}}
\expandafter\ifx\csname urlstyle\endcsname\relax
  \providecommand{\doi}[1]{doi: #1}\else
  \providecommand{\doi}{doi: \begingroup \urlstyle{rm}\Url}\fi

\bibitem[Chavdarova et~al.(2018)Chavdarova, Baqu{\'e}, Bouquet, Maksai, Jose,
  Bagautdinov, Lettry, Fua, Van~Gool, and Fleuret]{chavdarova2018wildtrack}
Tatjana Chavdarova, Pierre Baqu{\'e}, St{\'e}phane Bouquet, Andrii Maksai, Cijo
  Jose, Timur Bagautdinov, Louis Lettry, Pascal Fua, Luc Van~Gool, and Fran{\c
  c}ois Fleuret.
\newblock {WILDTRACK}: A multi-camera {HD} dataset for dense unscripted
  pedestrian detection.
\newblock In \emph{IEEE Conference on Computer Vision and Pattern Recognition
  (CVPR)}, pages 5030--5039, 2018.

\bibitem[Geifman and El-Yaniv(2019)]{geifman2019selectivenet}
Yonatan Geifman and Ran El-Yaniv.
\newblock {{SelectiveNet}: A Deep Neural Network with an Integrated Reject
  Option}.
\newblock In \emph{{Proceedings of the 36th International Conference on Machine
  Learning}}, pages 2151--2159, 2019.

\bibitem[Guo et~al.(2017)Guo, Pleiss, Sun, and Weinberger]{guo2017calibration}
Chuan Guo, Geoff Pleiss, Yu Sun, and Kilian~Q. Weinberger.
\newblock {On Calibration of Modern Neural Networks}.
\newblock In \emph{{Proceedings of the 34th International Conference on Machine
  Learning}}, pages 1321--1330, 2017.

\bibitem[Han et~al.(2023)Han, You, Wang, Zhang, Chu, Hu, Wang, and
  Liu]{han2023mmptrack}
Xiaotian Han, Quanzeng You, Chunyu Wang, Zhizheng Zhang, Peng Chu, Houdong Hu,
  Jiang Wang, and Zicheng Liu.
\newblock {MMPTRACK}: Large-scale densely annotated multi-camera multiple
  people tracking benchmark.
\newblock In \emph{IEEE/CVF Winter Conference on Applications of Computer
  Vision (WACV)}, pages 4849--4858, 2023.

\bibitem[Hao et~al.(2024)Hao, Liu, Zhan, Jin, Liu, Song, Hwang, and
  Wang]{hao2024divotrack}
Shengyu Hao, Peiyuan Liu, Yibing Zhan, Kaixun Jin, Zuozhu Liu, Mingli Song,
  Jenq-Neng Hwang, and Gaoang Wang.
\newblock {DIVOTrack: A Novel Dataset and Baseline Method for Cross-View
  Multi-Object Tracking in DIVerse Open Scenes}.
\newblock \emph{International Journal of Computer Vision}, 132\penalty0
  (4):\penalty0 1075--1090, 2024.

\bibitem[Luiten et~al.(2021)Luiten, O{\v s}ep, Dendorfer, Torr, Geiger,
  Leal-Taix{\'e}, and Leibe]{luiten2021hota}
Jonathon Luiten, Aljo{\v s}a O{\v s}ep, Patrick Dendorfer, Philip Torr, Andreas
  Geiger, Laura Leal-Taix{\'e}, and Bastian Leibe.
\newblock {HOTA}: A higher order metric for evaluating multi-object tracking.
\newblock \emph{International Journal of Computer Vision}, 129\penalty0
  (2):\penalty0 548--578, 2021.

\bibitem[Nguyen et~al.(2022)Nguyen, Henschel, Rosenhahn, Sonntag, and
  Swoboda]{nguyen2022lmgp}
Duy M.~H. Nguyen, Roberto Henschel, Bodo Rosenhahn, Daniel Sonntag, and Paul
  Swoboda.
\newblock {LMGP}: Lifted multicut meets geometry projections for multi-camera
  multi-object tracking.
\newblock In \emph{IEEE/CVF Conference on Computer Vision and Pattern
  Recognition (CVPR)}, pages 8856--8865, 2022.

\bibitem[Ristani et~al.(2016)Ristani, Solera, Zou, Cucchiara, and
  Tomasi]{ristani2016performance}
Ergys Ristani, Francesco Solera, Roger Zou, Rita Cucchiara, and Carlo Tomasi.
\newblock Performance measures and a data set for multi-target, multi-camera
  tracking.
\newblock In \emph{Computer Vision -- ECCV 2016 Workshops}, pages 17--35, 2016.

\bibitem[Specker(2024)]{specker2024ocmctrack}
Andreas Specker.
\newblock {OCMCTrack}: Online multi-target multi-camera tracking with
  corrective matching cascade.
\newblock In \emph{IEEE/CVF Conference on Computer Vision and Pattern
  Recognition Workshops (CVPRW)}, pages 7236--7244, 2024.

\bibitem[Woo et~al.(2024)Woo, Park, Shin, Kim, and Kweon]{woo2024mtmmc}
Sanghyun Woo, Kwanyong Park, Inkyu Shin, Myungchul Kim, and In~So Kweon.
\newblock {MTMMC}: A large-scale real-world multi-modal camera tracking
  benchmark.
\newblock In \emph{IEEE/CVF Conference on Computer Vision and Pattern
  Recognition (CVPR)}, pages 22335--22346, 2024.

\bibitem[Zhen et~al.(2026)Zhen, Xu, Wang, Fan, Dong, Zhao, and
  Fan]{zhen2026gmt}
Yihao Zhen, Mingyue Xu, Qiang Wang, Baojie Fan, Jiahua Dong, Tinghui Zhao, and
  Huijie Fan.
\newblock {GMT}: Effective global framework for multi-camera multi-target
  tracking.
\newblock In \emph{IEEE/CVF Conference on Computer Vision and Pattern
  Recognition (CVPR)}, pages 28201--28210, 2026.

\end{thebibliography}

\end{document}


\renewcommand{\thesection}{S\arabic{section}}
\renewcommand{\thetable}{S\arabic{table}}
\renewcommand{\thefigure}{S\arabic{figure}}
\renewcommand{\theequation}{S\arabic{equation}}
\twocolumn[{\centering
{\Large\bfseries CoWeave: Identity-Specific Context Reasoning\\for Persistent Multi-Camera Tracking}\par\medskip
{\large Supplementary Material}\par\bigskip}]
This supplement gives the evidence behind each claim of the main paper. For the fairness of the comparison, Section~\ref{sec:protocol-supp} defines the shared causal stream and the frozen arrival scoring, Table~\ref{tab:reference} compares CoWeave with trackers that use scene geometry or future tracklets, and Table~\ref{tab:gmt-protocol} shows that our GMT reproduction stays within $0.2$ CVIDF1 points of the published results under the original GMT protocol. For the data and uncertainty behind main-paper Table~1, Section~\ref{sec:sensitivity} reports identity counts, seed standard deviations, persistence, and cost on MTMMC, Table~\ref{tab:bootstrap} lists paired 95\% intervals on all four datasets, and Sections~\ref{app:additional} and~\ref{app:external-checked} give the full and per-seed results on WILDTRACK, MMPTrack, and DIVOTrack. For the contribution of each part, Section~\ref{sec:paired} crosses readers with matchers, varies the shared budget and the invalid-relation fraction, stratifies arrivals by evidence condition, and replaces single components, and Table~\ref{tab:sweeps} varies one setting at a time. Section~\ref{app:formal} completes the decision definitions, Section~\ref{app:operations} the online update loop and the paired replay of 100 ownership errors (Section~\ref{app:recovery}), and Sections~\ref{app:dataset-mtmmc}--\ref{app:dataset-divotrack} show the four datasets.

\section{Causal evaluation protocol}
\label{sec:protocol-supp}
To ensure that differences across methods come from cross-camera association and memory maintenance instead of detector quality, camera scheduling, or future lookahead, all complete-system comparisons share the causal observation stream and arrival cohort defined here. Sections~\ref{sec:sensitivity} and~\ref{sec:paired} report MTMMC operating points and controlled interventions under this shared protocol.

\paragraph{Observation and emission.} Every scheduled camera sample records three timestamps covering frame acquisition, completion of feature encoding, and output publication, and all three remain attached to every prediction. Detections, descriptors, candidate pools, and emission deadlines are fixed before the association rule changes, so a difference between pipelines cannot come from the detector or the camera schedule.

\paragraph{Arrival cohort.} In each camera, an arrival is a reference observation of a person who was absent from that camera's preceding scheduled observation after appearing earlier in another view. Because the person need not have exited the earlier view, overlapping cameras also produce arrivals. A person entering two cameras at the same decision step yields two arrivals in the cohort, and both observations enter a single joint decision in which compatible cross-camera views may share an identity.

To score each arrival without retrospective hindsight, we match boxes emitted before that step to reference annotations at $\mathrm{IoU}\geq0.5$ in their original camera and frame, compute a maximum-weight bipartite map over positively supported tracker-to-reference identity pairs, and freeze this map for the event. Continuing the person's mapped identity counts as correct. Assigning an established identity to a distractor counts as wrong, including steps where the tracker misses the arriving person while giving an established anchor to another person. Unassigned missed detections, new labels, unresolved outputs, and arrivals without a historical anchor form the new/unresolved class, partitioning all \Arrivals{} eligible MTMMC arrivals. Uninterrupted persistence $P_k$ requires a correct arrival followed by the same correct identity at each of the next $k$ eligible observations of that person.

\section{Identity metrics, persistence, and cost}
\label{sec:sensitivity}
Unlike the arrival map of Section~\ref{sec:protocol-supp}, which freezes past correspondences before each event, G-IDF1~\cite{ristani2016performance} fits one sequence-wide tracker-to-reference assignment over identity true positives (IDTP), false positives (IDFP), and false negatives (IDFN), combining precision $\mathrm{IDTP}/(\mathrm{IDTP}+\mathrm{IDFP})$ and recall $\mathrm{IDTP}/(\mathrm{IDTP}+\mathrm{IDFN})$ into $2\,\mathrm{IDTP}/(2\,\mathrm{IDTP}+\mathrm{IDFP}+\mathrm{IDFN})$. All pipelines share the reference detections, so $\mathrm{IDTP}+\mathrm{IDFN}$ is constant within each seed. Table~\ref{tab:ids}(a) shows that CoWeave raises both G-IDP and G-IDR, lifting G-IDF1 from \BaseID{} to \GoalID{} over causal GMT, and Table~\ref{tab:ids}(b) shows seed standard deviations of at most \SdMax{}, with wrong arrivals falling from \BaseW\% to \GoalW\% and new/unresolved outputs from \BaseN\% to \GoalN\%.

\begin{table*}[t]\centering
\caption{\textbf{Identity accounting and seed variation for the complete-system comparison on MTMMC.} (a) Mean identity counts over three training seeds under shared detections ($\mathrm{IDTP}+\mathrm{IDFN}$ fixed per seed), with precision, recall, and G-IDF1 in percent. (b) Means over three seeds with standard deviations in grey ($\le$\SdMax{} across all entries). In all supplementary tables, shading marks CoWeave, bold and underline mark the best and second best, $\Delta$ is the shaded row minus GMT (causal) unless stated otherwise, green and red $\Delta$ entries favor and disfavor the shaded row, and -- marks an inapplicable entry.}\label{tab:ids}
\small\setlength{\tabcolsep}{4pt}
\begin{tabularx}{\linewidth}{l*{6}{C}}\toprule
\rowcolor{tabgroup}\multicolumn{7}{l}{\textit{\textbf{(a) Identity accounting}}}\\
 & \multicolumn{3}{c}{\textbf{Mean identity counts}} & \multicolumn{3}{c}{\textbf{Identity scores}}\\
\cmidrule(lr){2-4}\cmidrule(lr){5-7}
\textbf{Method} & IDTP$\uparrow$ & IDFP$\downarrow$ & IDFN$\downarrow$ & G-IDP$\uparrow$ & G-IDR$\uparrow$ & G-IDF1$\uparrow$\\\midrule
\StyIdcounts
\bottomrule\end{tabularx}
\par\smallskip
\begin{tabularx}{\linewidth}{l*{7}{C}}\toprule
\rowcolor{tabgroup}\multicolumn{8}{l}{\textit{\textbf{(b) Means and seed standard deviations}}}\\
\textbf{Method} & G-IDF1$\uparrow$ & HOTA$\uparrow$ & Correct$\uparrow$ & Wrong$\downarrow$ & New/unres.$\downarrow$ & $P_{30}\uparrow$ & $R_{30}\uparrow$\\\midrule
\StySeeds
\bottomrule\end{tabularx}\end{table*}
Whereas G-IDF1 and HOTA, which averages $\sqrt{\mathrm{DetA}_a\mathrm{AssA}_a}$ across the 19 IoU thresholds $a=0.05,0.10,\ldots,0.95$~\cite{luiten2021hota}, evaluate association over whole sequences, uninterrupted persistence $P_k$ in Table~\ref{tab:profiles} isolates whether each arrival keeps its true identity across $k$ subsequent observations.

\begin{table*}[t]\centering
\caption{\textbf{Uninterrupted persistence across follow-up horizons on MTMMC.} $P_k$ is the percentage of arrivals with at least $k$ later observations that keep the correct identity at arrival and across all $k$ observations, from all \Arrivals{} arrivals at $P_0$ to the \Complete{} complete-follow-up arrivals at $P_{30}$. $R_{30}=P_{30}/P_0$ divides shares over nested cohorts and is therefore a retention ratio, not a survival probability.}\label{tab:profiles}
\small\setlength{\tabcolsep}{4pt}
\begin{tabularx}{\linewidth}{l*{6}{C}}\toprule
 & \multicolumn{5}{c}{\textbf{Uninterrupted persistence over $k$ later observations}} & \textbf{Ratio}\\
\cmidrule(lr){2-6}\cmidrule(lr){7-7}
\textbf{Method} & $P_0\uparrow$ & $P_5\uparrow$ & $P_{10}\uparrow$ & $P_{20}\uparrow$ & $P_{30}\uparrow$ & $R_{30}\uparrow$\\\midrule
\StyPersistent
\bottomrule\end{tabularx}\end{table*}
Across follow-up horizons in Table~\ref{tab:profiles}, CoWeave's lead over causal GMT widens from \LeadZero{} percentage points at arrival (\GoalC\% versus \BaseC\% at $P_0$) to \DeltaP{} points at $P_{30}$ (\GoalP\% versus \BaseP\%). Because the evaluated cohort shrinks from \Arrivals{} arrivals at $k=0$ to \Complete{} at $k=30$ as near-end arrivals lack $k$ subsequent observations, $R_{30}=P_{30}/P_0$ summarizes retention across nested cohorts rather than a survival probability over a fixed set.

\begin{table*}[t]\centering
\caption{\textbf{Runtime and storage cost under the common MTMMC workload.} Mean and P95 step latencies normalized to Independent${}=1$, absolute step durations in milliseconds with seed standard deviations in grey, and retained memory normalized to Independent${}=1$. Smaller values indicate lower online latency and smaller stored state.}\label{tab:stored}
\small\setlength{\tabcolsep}{4pt}
\begin{tabularx}{\linewidth}{l*{5}{C}}\toprule
 & \multicolumn{2}{c}{\textbf{Relative to Independent}} & \multicolumn{2}{c}{\textbf{Per-step duration (ms)}} & \textbf{Memory}\\
\cmidrule(lr){2-3}\cmidrule(lr){4-5}\cmidrule(lr){6-6}
\textbf{Method} & mean & P95 & mean & P95 & storage\\\midrule
\StyStored
\bottomrule\end{tabularx}\end{table*}
Table~\ref{tab:stored} verifies that these persistence gains come with bounded online overhead. Normalizing latency and retained bytes by the Independent baseline, CoWeave runs at a mean step time of \GoalMs{} ms, \TimeDelta\% above causal GMT (\BaseMs{} ms) and below MCTR (\MctrMs{} ms), while retaining \StoreSave\% less memory than Independent.

\begin{table*}[t]\centering
\caption{\textbf{Allocation of the shared \BudgetCap{}-unit evidence budget on MTMMC.} Each decision step has \BudgetCap{} units, shared by all active people, each spent on reading one unique source or on scoring one candidate. Entries report mean units per decision, mean step duration in milliseconds, and G-IDF1 (\%).}\label{tab:allocation}
\small\setlength{\tabcolsep}{4pt}
\begin{tabularx}{\linewidth}{l*{5}{C}}\toprule
 & \multicolumn{3}{c}{\textbf{Budget units per decision}} & \textbf{Online step} & \textbf{Quality}\\
\cmidrule(lr){2-4}\cmidrule(lr){5-5}\cmidrule(lr){6-6}
\textbf{Method} & unique reads & scorings & cap & mean ms & G-IDF1$\uparrow$\\\midrule
\StyCost
\bottomrule\end{tabularx}\end{table*}
Table~\ref{tab:allocation} shows how CoWeave spends the same \BudgetCap{}-unit cap. Relative to Independent, it moves \BudgetShift{} units per decision from reads to scorings, \CoScored{} against \InScored{}, and spends them on ranking sources and scoring each competing person.

\section{Component, budget, and relation analyses}
\label{sec:paired}
Every experiment in this section holds detections, descriptors, source availability, candidate generation, and emission times fixed at each prefix, so its differences come from the retrieval, scoring, and memory rules. Table~\ref{tab:fourcells} crosses two readers with two matchers under the shared \BudgetCap{}-unit budget: a tracklet reader ranks sources by similarity to the query, whereas the CoWeave reader ranks them by expected decision-loss reduction given the sources already read.

\begin{table*}[t]\centering
\caption{\textbf{Crossed reader and matcher comparison under a common MTMMC front end.} All four cells share identical observation prefixes, candidate sets, and the \BudgetCap{}-unit shared budget. $\Delta$ compares the full CoWeave pair against the tracklet reader with the GMT adapter. Scores are percentages, latency is in milliseconds, and storage is relative to Independent.}\label{tab:fourcells}
\small\setlength{\tabcolsep}{4pt}
\begin{tabularx}{\linewidth}{ll*{7}{C}}\toprule
 & & \textbf{Sequence} & \multicolumn{4}{c}{\textbf{Persistence}} & \multicolumn{2}{c}{\textbf{Cost}}\\
\cmidrule(lr){3-3}\cmidrule(lr){4-7}\cmidrule(lr){8-9}
\textbf{Reader} & \textbf{Matcher} & HOTA$\uparrow$ & $P_{10}\uparrow$ & $P_{20}\uparrow$ & $P_{30}\uparrow$ & $R_{30}\uparrow$ & ms & storage\\\midrule
\StyCrossed
\bottomrule\end{tabularx}\end{table*}
Starting from the tracklet reader with the GMT adapter in Table~\ref{tab:fourcells}, switching only the reader raises $P_{30}$ by \ReaderGain{} points without adding latency (\ReaderMs{} versus \TrackletMs{} ms), switching only the matcher raises $P_{30}$ by \MatcherGain{} points, and switching both raises $P_{30}$ by \BothGain{} points. The interaction, defined as the matcher gain under the CoWeave reader minus its gain under the tracklet reader, is \InteractionP{} points for $P_{30}$ with 95\% interval [\InteractionPlo, \InteractionPhi] and \Interaction{} points for G-IDF1 with interval [\InteractionLo, \InteractionHi] in Table~\ref{tab:bootstrap}. Both intervals include zero, so the test set shows no clear interaction: the reader and matcher gains add up.

\begin{table*}[t]\centering
\begin{minipage}[t]{0.493\linewidth}\centering
\caption{\textbf{G-IDF1 across shared read budgets on MTMMC.} Each budget cap covers all active people in a step. $\Delta$ is CoWeave minus GMT (causal) at the same cap, and the last column reports CoWeave's $P_{30}$. Means over three seeds (\%), with the default row matching Table~\ref{tab:ids}.}\label{tab:sbudget}
\footnotesize\setlength{\tabcolsep}{1.2pt}
\begin{tabularx}{\linewidth}{lCC>{\columncolor{tabours}}CCC}\toprule
 & \multicolumn{3}{c}{\textbf{G-IDF1$\uparrow$}} & & \textbf{CoWeave}\\
\cmidrule(lr){2-4}\cmidrule(lr){6-6}
\textbf{Budget} & Independent & GMT (causal) & CoWeave & $\Delta$ & $P_{30}\uparrow$\\\midrule
\StyBudget
\bottomrule\end{tabularx}\end{minipage}\hfill
\begin{minipage}[t]{0.493\linewidth}\centering
\caption{\textbf{G-IDF1 under controlled invalid-relation fractions on MTMMC.} Instance observations stay intact across rows while an increasing share of historical links is invalidated. $\Delta$ is CoWeave minus GMT (causal) at the same fraction, averaged over three seeds (\%).}\label{tab:svalidity}
\footnotesize\setlength{\tabcolsep}{1.5pt}
\begin{tabularx}{\linewidth}{lCC>{\columncolor{tabours}}CC}\toprule
 & \multicolumn{3}{c}{\textbf{G-IDF1$\uparrow$}} & \\
\cmidrule(lr){2-4}
\textbf{Invalid links} & Independent & GMT (causal) & CoWeave & $\Delta$\\\midrule
\StyValidity
\bottomrule\end{tabularx}\end{minipage}
\end{table*}
Tables~\ref{tab:sbudget} and~\ref{tab:svalidity} show that CoWeave holds its lead both when the shared read budget shrinks and when historical relations become invalid. Tightening the shared budget in Table~\ref{tab:sbudget} restricts how many sources a step can read, yet at one quarter of the default budget CoWeave reaches \QuarterCo{} G-IDF1 and \QuarterCoP\% $P_{30}$, matching causal GMT at twice the default budget (\DoubleGmt{} G-IDF1) and exceeding causal GMT's full-budget persistence (\BaseP\%). Corrupting historical relations at fixed source access in Table~\ref{tab:svalidity} instead tests whether stale context overrides intact instance cues. At 40\% invalid links, causal GMT loses \InvLossGmt{} G-IDF1 points and CoWeave \InvLossCo{}, as predicted relation validity lets valid instance evidence discount a broken link.

\begin{table}[t]\centering
\caption{\textbf{Arrival outcomes by evidence condition on MTMMC.} Correct, wrong, and new/unresolved shares (\%) partition all arrivals within each stratum, pooled over three training seeds, including the \SymArrivals{} completely symmetric arrivals across \SymEpisodes{} episodes.}\label{tab:strata}
\footnotesize\setlength{\tabcolsep}{2.5pt}
\begin{tabularx}{\linewidth}{l*{3}{C}}\toprule
\textbf{Condition and method} & Correct$\uparrow$ & Wrong$\downarrow$ & New/unres.$\downarrow$\\\midrule
\StyDiagnostics
\bottomrule\end{tabularx}\end{table}
Table~\ref{tab:strata} shows that CoWeave's arrival advantage grows where group context misleads causal GMT. Its lead in correct continuations over causal GMT widens from \LeadReliable{} points when relations are reliable and instance detail is clear, to \LeadGroup{} points under shared group context with weak instance detail, and to \LeadSplit{} and \LeadInvalid{} points after group separation and under invalid relations, lowering both wrong continuations and new/unresolved outputs in every asymmetric stratum. Under complete symmetry with equal priors, covering \SymArrivals{} arrivals across \SymEpisodes{} episodes where no accessible cue distinguishes the candidates at decision time, neither CoWeave nor causal GMT has separating evidence, so both leave all \SymArrivals{} arrivals unresolved with zero wrong continuations.

\begin{table*}[t]\centering
\caption{\textbf{Single-component replacements at the fixed \BudgetCap{}-unit budget on MTMMC.} Each row replaces one mechanism of CoWeave while keeping all other parts and the front end unchanged. Means over three training seeds (\%).}\label{tab:sablate}
\small\setlength{\tabcolsep}{4pt}
\begin{tabularx}{\linewidth}{l*{5}{C}}\toprule
\textbf{Configuration} & G-IDF1$\uparrow$ & Correct$\uparrow$ & Wrong$\downarrow$ & New/unres.$\downarrow$ & $P_{30}\uparrow$\\\midrule
\StyAblation
\bottomrule\end{tabularx}\end{table*}
Replacing any single component of CoWeave under the fixed \BudgetCap{}-unit budget in Table~\ref{tab:sablate} degrades both sequence and arrival accuracy. Fixed retrieval changes which records reach the matcher, removing relational support drops valid co-occurrence cues, freezing record ownership at insertion blocks later views from correcting misattributed history, and removing instance detail eliminates fine-grained appearance contrast, lowering G-IDF1 by \AblFixedRetID{} to \AblInstanceID{} points and $P_{30}$ by \AblFixedRet{} to \AblInstance{} points overall. In main-paper Table~3, reassigning whole tracklets at once under the same reader, budget, and emission rule reaches \RevGranoID{} G-IDF1, \RevGranoW\% wrong arrivals, and \RevGranoP\% $P_{30}$, confirming that record-level revision adds \RevGranID{} G-IDF1 and \RevGran{} $P_{30}$ points beyond tracklet-level correction.

\begin{table*}[t]\centering
\caption{\textbf{Memory retention policies and counterfactual relation interventions on MTMMC.} Means over three training seeds (\%) over the shared arrival cohort. (a) Retention policies under a common storage cap, reader, and matcher. (b) Relation configurations evaluated from identical snapshots, reads, and frozen decision pools.}\label{tab:retention-relations}
\small\setlength{\tabcolsep}{2pt}
\begin{tabularx}{\linewidth}{l*{5}{C}}\toprule
\textbf{Retention policy} & G-IDF1$\uparrow$ & Correct$\uparrow$ & Wrong$\downarrow$ & {\footnotesize New/unres.}$\downarrow$ & $P_{30}\uparrow$\\\midrule
\rowcolor{tabgroup}\multicolumn{6}{l}{\textit{\textbf{(a) Fixed reader, matcher, and storage capacity}}}\\
\StyRetention
\bottomrule\end{tabularx}
\par\smallskip
\begin{tabularx}{\linewidth}{l*{5}{C}}\toprule
\textbf{Relation configuration} & HOTA$\uparrow$ & $P_{10}\uparrow$ & $P_{20}\uparrow$ & $P_{30}\uparrow$ & $R_{30}\uparrow$\\\midrule
\rowcolor{tabgroup}\multicolumn{6}{l}{\textit{\textbf{(b) Fixed source contents, reads, and feasible decisions}}}\\
\StyIntervention
\bottomrule\end{tabularx}\end{table*}
Under a shared storage cap, reader, and matcher in Table~\ref{tab:retention-relations}(a), retaining records by predicted future decision risk per byte outperforms the alternative policies. When every policy receives identical incoming records and read budgets, CoWeave retention reaches \RetentionCo{} G-IDF1, beating the strongest G-IDF1 alternative (\RetBestID{} retention, \RetentionBest{}) by \RetentionMargin{} points and the strongest $P_{30}$ alternative (\RetBestP{} retention) by \RetMarginP{} points.

Counterfactual interventions on frozen snapshots in Table~\ref{tab:retention-relations}(b) confirm that relational gains come from linking the right identities rather than from link count alone. Starting every branch from identical stored records, ownership distributions, candidates, hypothesis beams, and the frozen decision pools of Section~\ref{app:operations}, the no-operation control reproduces the valid-relation row exactly. Degree-preserving permutation keeps each record's link count and time eligibility while rewiring which identities support each other, lowering $P_{30}$ by \PermCost{} points and matching the \ZeroCost{}-point drop from zeroing relational support altogether.

\section{Decision and evidence definitions}
\label{app:formal}

\begin{figure*}[t]
\centering
\input{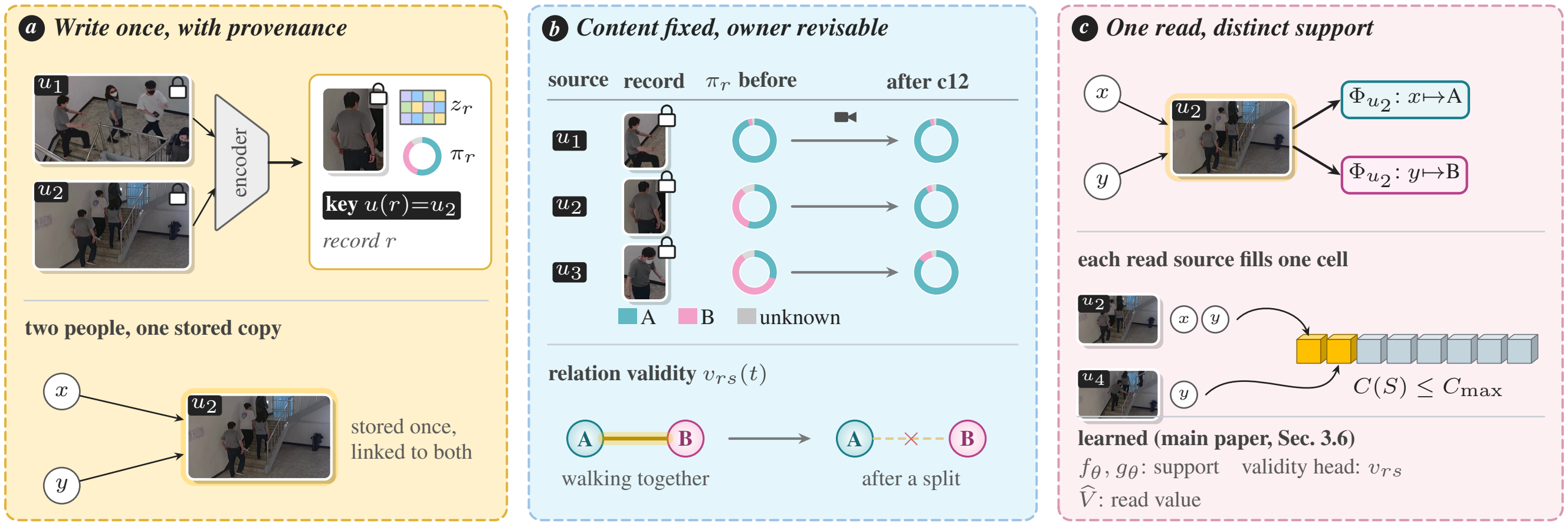}
\caption{\textbf{Immutable source storage, revisable ownership, and shared read accounting.} (a) Each historical source is encoded once into records storing its source key $u(r)$, descriptor $z_r$, and ownership distribution $\pi_r$, so co-occurring people share one stored source without duplication. (b) Raw record content remains locked (padlocks), whereas $\pi_r$ shifts across identities as later views arrive (before and after camera c12) and relation validity $v_{rs}(t)$ decays after a group splits. (c) Reading $u_2$ once supplies distinct candidate-conditioned support to $x$ and $y$ while charging a single unit toward $C(S)\le C_{\max}$. Identities (A${}={}$GT~113, B${}={}$GT~026), read set, and budget match the worked example of main-paper Figure~2.}
\label{fig:mechanism}
\label{fig:memory}
\end{figure*}
Figure~\ref{fig:mechanism} traces the worked example of main-paper Figure~2 across immutable source storage, post-emission ownership revision, and shared-budget reads. This section completes the constraints, fresh-identity naming, decision loss, and search limits behind that example.

\paragraph{Historical compatibility.}
When one frame contains several people, their records can contribute relational context but must never merge into a single person. Let $\mathcal I^{\rm hist}_t$ denote the pairs of historical records that cannot depict the same person after duplicate merging. Attention normalizes within each nonempty source $u$ so reading further sources leaves earlier weights unchanged, and the compatibility mask zeroes equal-identity relational terms on incompatible pairs:
\begin{equation}
\sum_{r\in\mathcal R_u}\alpha_{ir}=1,\qquad
\kappa_{rs}(j,k)=
\begin{cases}
0,&(r,s)\in\mathcal I^{\rm hist}_t,\ j=k,\\
1,&\text{otherwise}.
\end{cases}
\end{equation}
Here $\alpha_{ir}\geq0$, an empty source contributes zero, and $\kappa_{rs}(j,j)=0$ excludes an incompatible pair even when both of its ownership marginals assign positive mass to identity $j$.

\paragraph{Fresh identities and unresolved actions.}
Comparing joint hypotheses that introduce new people requires removing label-permutation duplicates and decomposing search across unrelated arrivals. For $n_t$ current observations, fresh labels $\mathcal B_t=\{\nu_1,\ldots,\nu_{n_t}\}$ are disjoint from established identities $\mathcal J_t$, and every fresh-identity group is named canonically by its smallest observation index:
\begin{equation}
H(i)=\nu_b\ \Longrightarrow\ b=\min\{k:H(k)=\nu_b\}.
\end{equation}
Restricting $\mathcal H_t$ to canonical names prevents relabeled duplicates from inflating new-identity probability mass. Emitted decisions obey the same feasibility rules on committed labels, while unresolved outputs ($\bot$) carry no shared identity label. A true identity absent from the candidate set counts as a retrieval miss and is never supervised as a fresh birth.

To allow independent solving across components, ambiguity components include an edge for every observation pair permitted to share a fresh label alongside shared historical candidates, incompatibilities, and active relations. Because these fresh-group edges keep every admissible multi-view birth within one component and canonical names use global observation indices, fresh labels never cross component boundaries. The relabeling match $\sigma$ in Eq.~\ref{eq:explicit-loss} therefore separates across components because groups in distinct components share no observation, combining local optimal relabelings into the exact global loss so two views of one new person cannot split into two zero-cost births.

\paragraph{Read-value regression target.}
To select sources that change emitted decisions, the reader is supervised by the actual loss drop under the online solver. On each training prefix, adding an eligible source $u$ yields a target equal to the realized decrease in Eq.~\ref{eq:explicit-loss} under identical candidates, action costs, and beam widths, while candidate-expansion actions are charged separately and shared across both sides of subsequent read comparisons. Squared-loss regression estimates the conditional mean of this drop across training episodes. With exact conditional posteriors and minimum-risk decisions, the tower property makes the expected value of information nonnegative, but finite beam search can produce negative realized drops on individual steps, so online retrieval reads only sources with positive predicted value.

\paragraph{Shared decision loss.}
Let $d_i=D(i)$ and $h_i=H(i)$ denote the emitted action and hypothesized label of observation $i$, with established identities in $\mathcal J_t$ compared directly and fresh groups in $\mathcal B_t$ matched one-to-one by $\sigma$ using distinct dummy labels for unmatched groups. The identity decision loss minimizes over $\sigma$:
\begin{equation}
\label{eq:explicit-loss}
\begin{gathered}
L(D,H)=\min_{\sigma}\sum_i\ell(d_i,h_i;\sigma),\\
\ell(d_i,h_i;\sigma)=
\begin{cases}
\lambda_U,&d_i=\bot,\\
\lambda_W,&d_i\in\mathcal J_t,\ d_i\ne h_i,\\
\lambda_B,&d_i\in\mathcal B_t,\ \sigma(d_i)\ne h_i,\\
0,&\text{otherwise}.
\end{cases}
\end{gathered}
\end{equation}
Setting $\lambda_W=1$ for a wrong continuation, $\lambda_B=0.5$ for an unmatched fresh identity, and $\lambda_U=0.35$ for an unresolved output aligns decision emission, read-value supervision, and retention risk under one cost function.

\paragraph{Hypothesis and decision searches.}
All matched comparisons share three search limits that keep joint inference within the step budget. Each observation retains its 32 highest-scoring admissible historical candidates plus canonical fresh-group actions. A width-16 hypothesis beam then extends observations in canonical order, enforces incompatibilities, scores completed factors, and breaks ties canonically. Scores are normalized over retained complete hypotheses as $P_S(H)\propto\exp(F(H;S)/T)$, with temperature $T>0$ fitted by development-set negative log-likelihood on covered examples and frozen before testing, while uncovered arrivals remain in full-cohort evaluation as search-coverage misses. Table~\ref{tab:coverage} checks whether these finite limits bottleneck accuracy.

\begin{table}[t]\centering
\caption{\textbf{Candidate and hypothesis-beam coverage over evaluable MTMMC arrivals.} Candidate coverage is the share of arrivals whose true identity survives the 32-candidate cap, and correct-hypothesis coverage is the share whose true joint assignment survives the width-16 beam. All methods start from the same generated candidates, and coverage counts those each method retains for its search. Full-cohort G-IDF1 is shown alongside (\%).}\label{tab:coverage}
\footnotesize\setlength{\tabcolsep}{2.5pt}
\begin{tabularx}{\linewidth}{l*{3}{C}}\toprule
 & \multicolumn{2}{c}{\textbf{Search coverage}} & \textbf{Outcome}\\
\cmidrule(lr){2-3}\cmidrule(lr){4-4}
\textbf{Configuration} & Candidate$\uparrow$ & Correct hypothesis$\uparrow$ & Full-cohort G-IDF1$\uparrow$\\\midrule
\StyCoverage
\bottomrule\end{tabularx}\end{table}
In Table~\ref{tab:coverage}, CoWeave keeps the true identity among its 32 retained candidates at 96.9\% of evaluable arrivals and preserves the true joint assignment inside its width-16 beam at 95.1\%, compared with 92.0\% and 86.9\% for causal GMT from the same initial candidate pool. At the same search limits, fewer CoWeave arrivals are lost to the finite search.

Given $P_S(H)$, a width-8 decision beam extends partial decisions with admissible established IDs, canonical fresh groups, or unresolved outputs ($\bot$), ranking partial prefixes by posterior expected loss with unassigned observations treated as unresolved. The lowest-risk completed decision in the beam defines $D_S$ before and after every read, and the same decoder estimates future query risk during memory retention in Eq.~\ref{eq:retention}.

\section{Online updates, computation, and paired replay}
\label{app:operations}
Because the source bank stores each historical observation once and never rewrites its content, revising ownership $\pi_r$ or relation validity $v_{rs}(t)$ changes only how stored evidence is weighted at later arrivals, never past sensor payloads or already emitted identities. This section gives the online update loop, its cost, training, and the counterfactual controls, and ends with the paired replay of 100 ownership errors cited in the main paper (Section~\ref{app:recovery}).

\subsection{State and update order}
Keeping source payloads, instance ownership, context validity, and decision logs separate allows error auditing across four distinct causes, namely unavailable evidence, candidate truncation, uncertain ownership, and stale relations. Specifically, a source key binds raw content to its representation version, an instance record stores the observation key, descriptor, candidate support, and $\pi_r$, a context record stores supporting instances, relation type, validity $v_{rs}(t)$, and update time, and an emitted decision logs its action, candidate set, source dependencies, and emission time.

Every online decision step executes eight operations in order and emits its output before mutating memory state:
\begin{enumerate}
\item Ingest the current observations and all historical sources encoded by the decision time.
\item Generate candidates from resident identity summaries and the admissible source directory, logging candidate truncation separately from assignment decisions.
\item Form ambiguity components from shared historical candidates, incompatibilities, active relations, and admissible fresh-identity pairs using global canonical fresh labels, and run the width-16 hypothesis beam of Section~\ref{app:formal} with $\pi_r$ and $v_{rs}(t)$ fixed for the whole read loop of this step.
\item Normalize hypothesis posteriors, run the width-8 decision beam over committed and unresolved actions under Eq.~\ref{eq:explicit-loss}, enumerate accessible unread sources, and charge directory expansion to the shared budget.
\item Rank unread sources by predicted loss reduction per incremental cost, skipping any source that exceeds the remaining budget or becomes available after the deadline.
\item Read the top-ranked source once, route its records to all affected components, update hypothesis posteriors and minimum-loss decisions within the same beam limits, and rescore remaining unread sources while keeping $\pi_r$ fixed within the step.
\item Terminate retrieval once no admissible source has positive predicted value, the budget is exhausted, or the deadline arrives, and emit the lowest-loss completed decision in the beam, including any unresolved outputs.
\item Revise ownership and relation validity after emission. New records take the posterior marginal of main-paper Eq.~7, re-read historical records recompute $\pi_r$ from their deduplicated factor logs, dependent resident summaries are refreshed so future base scores never double-count revised records, the storage cap is enforced, and the decision is logged without relabeling earlier outputs.
\end{enumerate}

Because all active people draw from one shared step budget, reading a multi-person source once serves every person in it at a single read cost. Similarly, tagging each logged factor with its identifier and representation version lets $\pi_r$ be recomputed after updates with each factor counted once.

Step~8 bounds memory within $M$ bytes, where $\operatorname{bytes}(A)$ sums unique sources, keys, descriptors, ownership states, retained relations, and indexes across retained subset $A$, by minimizing expected future decision risk:
\begin{equation}
\label{eq:retention}
A^*=\arg\min_{A:\operatorname{bytes}(A)\leq M}
\mathbb E_{Q\sim\widehat p_t}
  [\rho_Q(A)].
\end{equation}
Here $\rho_Q(A)$ is the model-estimated minimum decision risk on query $Q$ using retained state $A$, and $\widehat p_t$ predicts upcoming identity queries from training statistics and causal history without accessing future test frames. Rather than keeping the most recent or most similar records, greedy pruning evicts the source bundle with the smallest predicted risk increase per byte recovered, removing its dependent records together and rescoring remaining bundles after each eviction.

\subsection{Computational cost}
For an ambiguity component of $n$ observations with at most $k$ candidates each, $N_H$ retained hypotheses, and $r$ records exposed by a read source, scoring instance terms takes $O(N_Hnr)$ operations and relational pair terms take up to $O(N_Hn^2r^2)$. Hypothesis-independent compatibility values are cached within the step and charged in the budget log.

Because pairwise relational terms prevent linear-assignment reduction, small components use exact enumeration whereas larger ones use the width-16 beam under identical feasibility constraints, logging search truncation separately from scoring errors. Storage decomposes into $M_{\rm source}+M_{\rm instance}+M_{\rm ownership}+M_{\rm relation}+M_{\rm index}$, and refreshing summaries from the factor log in a single pass avoids multiplying an observation repeatedly across revisions. Reported mean and P95 step latencies include all work inside the online boundary, covering feature extraction or decoding where applicable, directory lookup, read-value selection, source retrieval, joint scoring, and post-emission state updates.

\subsection{Learning and calibration}
Training episodes run on causal observation prefixes so ground-truth identities supervise assignment likelihood and read-value targets only after candidate sets are constructed. Candidate misses and beam-pruned true assignments mask the assignment-likelihood term during training while remaining in test-time full-cohort scoring, and fresh-label supervision applies only to genuine first appearances.

Ownership and relation validity receive separate supervision because co-walking people can form a valid group while their individual identities remain uncertain, and a subsequent split invalidates their group link without altering who appeared in earlier views. Ambiguous relation pairs are masked during training, and calibration temperature $T$ is fitted on disjoint development sequences under fixed decision costs, following standard confidence calibration~\cite{guo2017calibration} and selective prediction~\cite{geifman2019selectivenet}.

\subsection{Counterfactual controls and paired intervals}
To prevent repeated reads of one source from double-counting its evidence during ownership updates, provenance keys keep base summaries and retrieved source details in disjoint feature slots. For memory state $\mathcal M_t$ and factor-update operator $U$, replaying an active factor $e$ with the same identifier and version is idempotent:
\begin{equation}
U(U(\mathcal M_t,e),e)=U(\mathcal M_t,e).
\end{equation}
Re-accessing the source still incurs its online read cost, while $U$ rebuilds $\pi_r$ from the deduplicated factor log instead of treating a prior posterior as fresh evidence, and confirmed split relations remain negative examples for validity training.

This deterministic state log supports the counterfactual branches of Table~\ref{tab:retention-relations}. Replaying the unmodified tracker reproduces its exact output stream and cost log. In Table~\ref{tab:retention-relations}(b), all relation branches fork from the same causal snapshot at each evaluated step with identical source contents, read paths, ownership, ambiguity components, candidates, and width-16 hypothesis beams. Degree-preserving permutation rewires relational links while preserving node and edge counts, degree sequence, and time eligibility (retaining snapshots that lack a nontrivial valid permutation), and zeroing sets relational weights to zero along the identical read path.

Freezing only the hypothesis beam would still let altered relational scores change which partial decisions survive the width-8 decision beam. Before comparing relation branches at a snapshot, we therefore run each branch's decision beam on the shared inputs, take the union of their feasible completed decisions plus the all-unresolved and all-fresh-identity actions, and freeze this common pool without ground truth or future frames. Scoring every branch over the identical pool under Eq.~\ref{eq:explicit-loss} with canonical tie-breaking rules out decision-beam pruning as a confound, with shared pool construction and branch-specific rescoring logged under the declared budget. In contrast, end-to-end ablations run independent branch streams, and Table~\ref{tab:retention-relations}(a) replays one fixed query stream across retention policies so a difference between policies cannot come from the reader.

For the MTMMC intervals in Table~\ref{tab:bootstrap}, we draw 10,000 paired bootstrap resamples over the \Episodes{} test episodes with shared episode keys across methods, recomputing global sufficient statistics, nonlinear G-IDF1 and HOTA, and the four-cell reader--matcher interaction within each resample. Thresholds and stopping rules are set on development data only. Decision and sequence logs also attribute each error to an evicted distinguishing detail, a pruned true candidate, misscored accessible evidence, or a relation that survives a group split.

\subsection{Paired replay after an ownership error}
\label{app:recovery}
Uninterrupted persistence $P_k$ assigns zero to any arrival that starts wrong, so it cannot reveal whether a tracker stays trapped in an identity swap or recovers once clearer views arrive. The paired replay in Table~\ref{tab:repair} measures post-error recovery directly. Each replay event starts two branches from the identical state snapshot of the common reference tracker right after an erroneous ownership update, sharing the exact historical source bank, mistaken ownership state, and future observation schedule. Identity anchors are fixed from the prefix before the error, and ground truth is used only by the evaluator to check whether the original anchored identity returns, never exposed to either tracker branch.

Both branches run the same reader, shared budget, source bank, and relation-validity update rule, differing solely in whether pre-error ownership can be revised. The frozen branch locks existing ownership distributions at the erroneous snapshot while assigning insertion-time marginals to later records (unlike the fixed-ownership ablation of Table~\ref{tab:sablate}, which freezes every record from sequence start), whereas the revisable branch recomputes $\pi_r$ whenever a source is re-read, and a no-operation check confirms identical starting states. A recovery counts only when the tracker emits a continuation of the original anchored identity, so following a distractor, shifting boxes, assigning a fresh label, or abstaining does not count, and missed predictions at horizon $h$ count as new/unresolved.

Across 100 erroneous-ownership events with complete follow-up to $h=30$ subsequent observations in Table~\ref{tab:repair}(a), revisable ownership restores the original identity at least once in \RepairEver{} cases after a median of \RepairMedian{} steps, and \RepairHeld{} cases (\RepairHeldPct{}\% of those recovered) retain it at $h=30$, while \RepairLost{} lose it again before $h=30$ and \RepairNever{} never return. Neither frozen ownership nor its no-operation check recovers any event. Table~\ref{tab:repair}(b) evaluates revision safety on 200 snapshots that start with correct ownership under the same two branches, where revision breaks \pv{2} continuations (\pv{1.0}\%) and leaves all other outcomes unchanged.

\begin{table*}[t]\centering
\caption{\textbf{Paired replay of post-error recovery and revision safety at horizon $h=30$.} Both branches start from identical snapshots and process identical future observations under the same reader, budget, and relation-validity rule, differing only in whether ownership is revisable. (a) 100 snapshots starting from an ownership error (counts equal percentages). (b) 200 snapshots starting with correct ownership, where Newly broken counts continuations that revision breaks against the frozen branch. $\Delta$ is revisable minus frozen ownership.}\label{tab:repair}
\small\setlength{\tabcolsep}{4pt}
\begin{tabularx}{\linewidth}{l*{4}{C}}\toprule
\textbf{Branch} & Original identity$\uparrow$ & Wrong identity$\downarrow$ & New/unresolved$\downarrow$ & Newly broken$\downarrow$\\\midrule
\StyRepair
\bottomrule\end{tabularx}\end{table*}

\section{Training, intervals, transfer, and sensitivity}
\label{app:additional}
Table~\ref{tab:training} lists the search settings shared by the three training seeds and the training curves of the relation-validity loss, the read-value ranking loss, and development G-IDF1.

\begin{table*}[t]\centering
\caption{\textbf{Shared solver settings and multi-seed training progression.} Left: default search limits and unresolved cost $\lambda_U$ from Section~\ref{app:formal}, shared by all three training seeds, with one checkpoint per seed and no overlap between training, development, and test sequences. Right: relation-validity binary cross-entropy (BCE), read-value ranking loss, and development G-IDF1 (\%) across training progress, shown as three-seed means with standard deviations in grey.}\label{tab:training}
\small\begin{tabular*}{.3\linewidth}[t]{@{\extracolsep{\fill}}lr@{}}\toprule \textbf{Setting} & \textbf{Value}\\\midrule
Independent training seeds & \pv{3}\\
Candidate limit & \pv{32}\\
Hypothesis beam & \pv{16}\\
Decision beam & \pv{8}\\
Unresolved cost $\lambda_U$ & \pv{0.35}\\
\bottomrule\end{tabular*}\hfill
\begin{tabular*}{.64\linewidth}[t]{@{\extracolsep{\fill}}cccc@{}}\toprule \textbf{Progress (\%)} & Validity BCE$\downarrow$ & Ranking loss$\downarrow$ & Dev.\ G-IDF1$\uparrow$\\\midrule\StyTraining\bottomrule\end{tabular*}
\end{table*}

\begin{table*}[t]\centering
\caption{\textbf{Paired 95\% confidence intervals for main CoWeave gains and reader--matcher interactions.} MTMMC intervals use 10,000 paired bootstrap resamples over the \Episodes{} test episodes, recomputing each interaction from its four cells. External G-IDF1 intervals on WILDTRACK, MMPTrack, and DIVOTrack use paired differences across the three training seeds ($\bar d\pm t_{0.975,2}s_d/\sqrt{3}$). Entries are percentage points computed from unrounded means, which may differ by 0.1 from differences of rounded table entries.}\label{tab:bootstrap}
\small\setlength{\tabcolsep}{4pt}
\begin{tabularx}{0.8\linewidth}{l*{3}{C}}\toprule
\textbf{Quantity} & Estimate & 95\% interval & Excludes zero\\\midrule
\StyBootstrap
\bottomrule\end{tabularx}
\end{table*}
In Table~\ref{tab:bootstrap}, every interval for a difference between CoWeave and causal GMT excludes zero, on MTMMC and on the three other datasets, whereas both reader--matcher interaction intervals include zero. On WILDTRACK, MMPTrack, and DIVOTrack, the intervals use the three paired seed differences with two degrees of freedom.

\begin{table}[t]\centering
\caption{\textbf{Comparison on MTMMC with trackers that use inputs beyond the causal RGB stream.} CoWeave uses only the causal RGB stream, whereas the reference methods add scene geometry or future tracklets. All scores are G-IDF1 on the MTMMC test episodes under the protocol of Section~\ref{sec:protocol-supp}; the citations identify the methods.}\label{tab:reference}
\footnotesize\setlength{\tabcolsep}{2.5pt}
\begin{tabularx}{\linewidth}{llC}\toprule
\textbf{Method} & \textbf{Input} & G-IDF1$\uparrow$\\\midrule
\StyReference
\bottomrule\end{tabularx}
\end{table}
In Table~\ref{tab:reference}, CoWeave (\GoalID{} G-IDF1 from the causal RGB stream alone) surpasses both MTMMC reference trackers that access richer inputs, OCMCTrack~\cite{specker2024ocmctrack} with scene geometry and lifted multicuts~\cite{nguyen2022lmgp} with future tracklets.

\begin{table*}[t]\centering
\caption{\textbf{Full metric comparison on WILDTRACK, MMPTrack, and DIVOTrack under the unchanged MTMMC protocol.} Entries report three-seed means (\%), mean online step duration in milliseconds, and $\Delta$ as CoWeave minus GMT (causal). $R_{30}=P_{30}/P_0$, where $P_0$ equals the Correct arrival share.}\label{tab:transfer}
\small\setlength{\tabcolsep}{4pt}
\begin{tabularx}{\linewidth}{l*{8}{C}}\toprule
\textbf{Dataset and method} & G-IDF1$\uparrow$ & HOTA$\uparrow$ & Correct$\uparrow$ & Wrong$\downarrow$ & {\footnotesize New/unres.}$\downarrow$ & $P_{30}\uparrow$ & $R_{30}\uparrow$ & ms\\\midrule
\StyTransfer
\bottomrule\end{tabularx}
\end{table*}
Across WILDTRACK, MMPTrack, and DIVOTrack in Table~\ref{tab:transfer}, CoWeave leads all five baselines on every accuracy metric under the unchanged MTMMC protocol, with scene views in Sections~\ref{app:dataset-wildtrack}--\ref{app:dataset-divotrack} and per-seed counts in Section~\ref{app:external-checked}. Over causal GMT, CoWeave raises G-IDF1 by \TransferMin{} to \TransferMax{} points, HOTA by \TransHMin{} to \TransHMax{} points, and $P_{30}$ by \TransferPMin{} to \TransferPMax{} points while lowering wrong arrivals to \TransWcMin\% to \TransWcMax\%, about half of causal GMT's \TransWgMin\% to \TransWgMax\%.

\begin{table*}[t]\centering
\caption{\textbf{One-factor hyperparameter sensitivity of CoWeave on MTMMC.} Each row varies one decision or search hyperparameter while holding all others at their defaults (shaded), reporting setting value: G-IDF1\,/\,$P_{30}$ (\%).}\label{tab:sweeps}
\small\setlength{\tabcolsep}{3pt}
\begin{tabularx}{\linewidth}{l*{5}{C}}\toprule
\textbf{Setting} & \multicolumn{5}{c}{\textbf{Value: G-IDF1\,/\,$P_{30}$}}\\\midrule
\StySweeps
\bottomrule\end{tabularx}
\end{table*}
Single-factor sweeps across all decision and search hyperparameters in Table~\ref{tab:sweeps} keep CoWeave at or above \SweepMinID{} G-IDF1 and \SweepMinP\% $P_{30}$ on MTMMC, above causal GMT's default \BaseID{} G-IDF1 and \BaseP\% $P_{30}$. Varying unresolved cost $\lambda_U$, fresh-label cost $\lambda_B$, relational weight $\lambda_{\mathrm{rel}}$, temperature $T$, candidate cap, and beam widths one at a time gives the lowest scores at the lowest $\lambda_{\mathrm{rel}}$ and the tightest candidate cap, which restrict the evidence that separates competing people.

\section{External per-seed results and GMT protocol check}
\label{app:external-checked}
Across WILDTRACK, MMPTrack, and DIVOTrack in Tables~\ref{tab:external-sd}--\ref{tab:external-counts}, CoWeave outperforms causal GMT on every training seed over fixed arrival cohorts, with sample standard deviations at or below \SdMax{}. Specifically, Table~\ref{tab:external-sd} reports the three-seed means and standard deviations behind main-paper Table~1 and Table~\ref{tab:transfer}, Table~\ref{tab:external-seeds} lists G-IDF1 and $P_{30}$ for each paired seed, and Table~\ref{tab:external-counts} provides the fixed arrival denominators ($N_{30}=984$, $5{,}136$, and $3{,}012$) and integer success counts, for mean gains over causal GMT of \TransferMin{}--\TransferMax{} G-IDF1 and \TransferPMin{}--\TransferPMax{} $P_{30}$ points.

To check that causal GMT's scores in main-paper Table~1 do not stem from a weak reproduction, Table~\ref{tab:gmt-protocol} evaluates our GMT implementation under the original GMT protocol and CVIDF1 metric~\cite{zhen2026gmt}. Across three runs, reproduced scores stay within $0.2$ points of the published numbers ($71.8\pm0.3$ versus $72.0$ on WILDTRACK and $73.3\pm0.4$ versus $73.2$ on DIVOTrack), so the lower GMT scores in main-paper Table~1 reflect the causal shared-budget setting rather than the reproduction.

\begin{table}[t]\centering
\caption{\textbf{Means and sample standard deviations on WILDTRACK, MMPTrack, and DIVOTrack.} Three paired training seeds (\%), with standard deviations at or below \SdMax{}. Each $P_{30}$ is evaluated over the fixed arrival denominator $N_{30}$ of Table~\ref{tab:external-counts}, shared across methods and seeds.}\label{tab:external-sd}
\footnotesize\setlength{\tabcolsep}{2.5pt}
\begin{tabularx}{\linewidth}{llCC}\toprule
\textbf{Dataset} & \textbf{Method} & G-IDF1 (mean $\pm$ SD) & $P_{30}$ (mean $\pm$ SD)\\\midrule
WILDTRACK & GMT (causal) & 70.2 $\pm$ 0.36 & 63.5 $\pm$ 0.41 \\
WILDTRACK & CoWeave & 74.8 $\pm$ 0.44 & 73.2 $\pm$ 0.53 \\
MMPTrack & GMT (causal) & 66.6 $\pm$ 0.46 & 59.9 $\pm$ 0.46 \\
MMPTrack & CoWeave & 74.0 $\pm$ 0.61 & 75.5 $\pm$ 0.52 \\
DIVOTrack & GMT (causal) & 63.1 $\pm$ 0.36 & 56.0 $\pm$ 0.36 \\
DIVOTrack & CoWeave & 69.3 $\pm$ 0.44 & 68.3 $\pm$ 0.54 \\
\bottomrule\end{tabularx}\end{table}

\begin{table*}[t]\centering
\caption{\textbf{Per-seed G-IDF1 and $P_{30}$ on WILDTRACK, MMPTrack, and DIVOTrack.} Each cell reports G-IDF1 / $P_{30}$ (\%) for one of the three paired training seeds.}\label{tab:external-seeds}
\small\setlength{\tabcolsep}{4pt}
\begin{tabularx}{0.8\linewidth}{llCCC}\toprule
\textbf{Dataset} & \textbf{Method} & Seed 1 & Seed 2 & Seed 3\\\midrule
WILDTRACK & GMT (causal) & 70.1 / 63.11 & 70.6 / 63.92 & 69.9 / 63.52 \\
WILDTRACK & CoWeave & 74.3 / 72.56 & 75.1 / 73.48 & 75.0 / 73.48 \\
MMPTrack & GMT (causal) & 66.2 / 59.50 & 67.1 / 60.40 & 66.5 / 59.79 \\
MMPTrack & CoWeave & 73.6 / 75.19 & 74.7 / 76.09 & 73.7 / 75.19 \\
DIVOTrack & GMT (causal) & 62.8 / 55.61 & 63.5 / 56.11 & 63.0 / 56.31 \\
DIVOTrack & CoWeave & 69.0 / 67.73 & 69.1 / 68.79 & 69.8 / 68.39 \\
\bottomrule\end{tabularx}\end{table*}

\begin{table*}[t]\centering
\caption{\textbf{Fixed arrival denominators and per-seed persistence counts on external datasets.} A success requires the correct identity at arrival and throughout all 30 subsequent observations. Denominator $N_{30}$ is fixed across methods and seeds, counting missed detections and unresolved outputs as failures, with $P_{30}=100\times N_{\rm success}/N_{30}$.}\label{tab:external-counts}
\small\setlength{\tabcolsep}{4pt}
\begin{tabularx}{0.8\linewidth}{lCCCC}\toprule
\textbf{Dataset} & All arrivals & $N_{30}$ & \shortstack{GMT successes\\seeds 1 / 2 / 3} & \shortstack{CoWeave successes\\seeds 1 / 2 / 3}\\\midrule
WILDTRACK & 1,264 & 984 & 621 / 629 / 625 & 714 / 723 / 723 \\
MMPTrack & 6,487 & 5,136 & 3,056 / 3,102 / 3,071 & 3,862 / 3,908 / 3,862 \\
DIVOTrack & 3,916 & 3,012 & 1,675 / 1,690 / 1,696 & 2,040 / 2,072 / 2,060 \\
\bottomrule\end{tabularx}\end{table*}

\begin{table*}[t]\centering
\caption{\textbf{GMT fidelity check under its original evaluation protocol.} Published CVIDF1 values come from GMT~\cite{zhen2026gmt} (Table~2), and reproduced values report means and sample standard deviations over three runs. $\Delta$ is reproduced minus published CVIDF1 in percentage points. This protocol check is separate from the causal shared-budget G-IDF1 comparison in Table~\ref{tab:transfer}.}\label{tab:gmt-protocol}
\small\setlength{\tabcolsep}{4pt}
\begin{tabularx}{0.8\linewidth}{lCCCC}\toprule
\textbf{Dataset} & Published CVIDF1 & Reproduced CVIDF1 & Runs 1 / 2 / 3 & $\Delta$\\\midrule
WILDTRACK & 72.0 & $71.8\pm0.3$ & 71.5 / 71.9 / 72.0 & $-0.2$ \\
DIVOTrack & 73.2 & $73.3\pm0.4$ & 73.0 / 73.2 / 73.7 & $+0.1$ \\
\bottomrule\end{tabularx}\end{table*}

\section{Cross-camera views and occlusions in MTMMC}
\label{app:dataset-mtmmc}
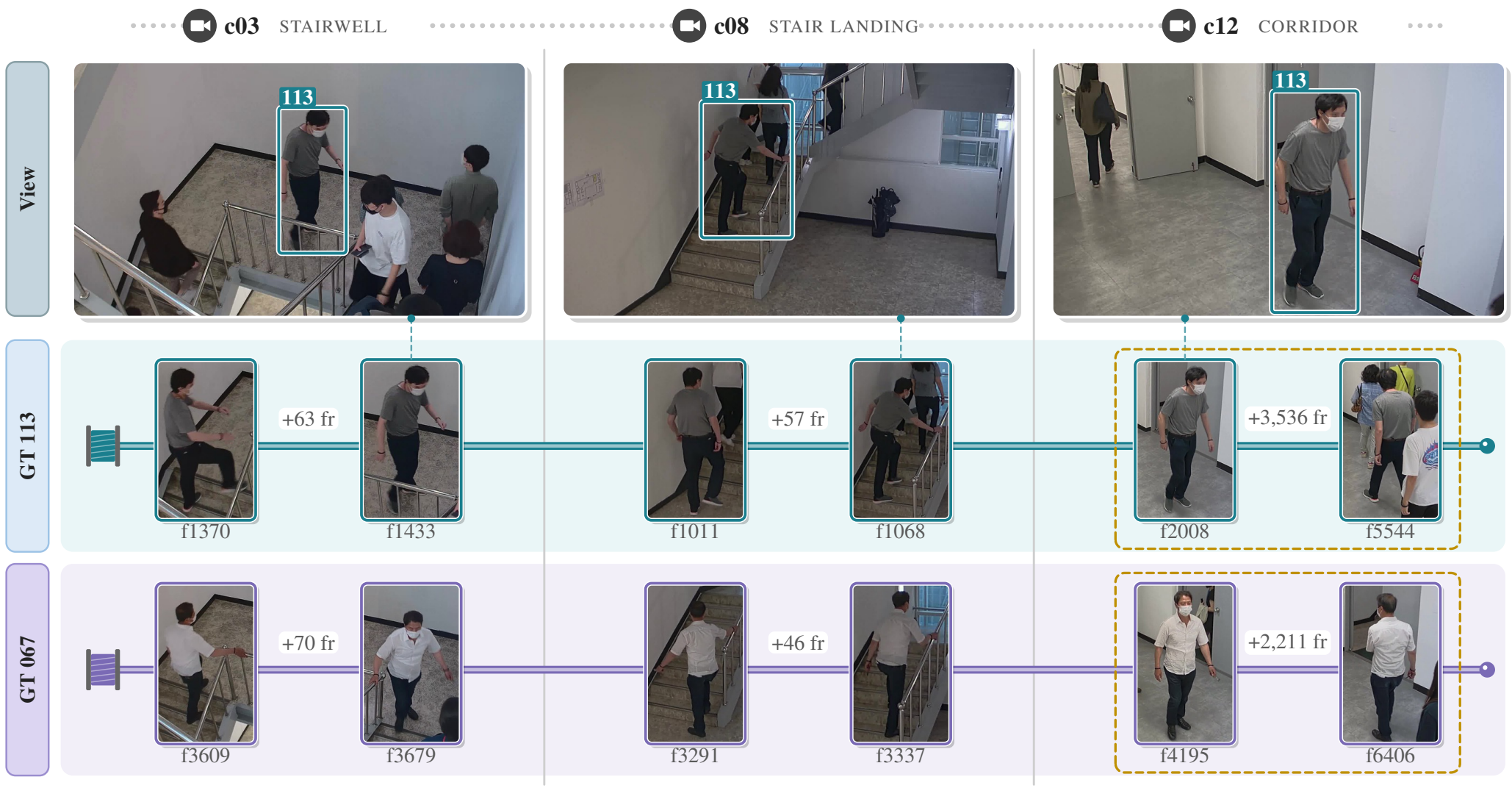
\begin{figure*}[t]\centering
\resizebox{0.78\linewidth}{!}{\begin{tikzpicture}[cw,font=\footnotesize]
\def\W{496}
\path (0,40) (\W,300);
\def\CW{150}\def\CG{12}\def\XL{22}
\def\Xa{22}\def\Xb{184}\def\Xc{346}
\draw[cwInk!25,line width=1.6pt,line cap=round,dash pattern=on 0pt off 3.2pt] (\Xa+20,292) -- (\Xc+130,292);
\foreach \X/\cam/\place in {\Xa/c03/stairwell,\Xb/c08/stair landing,\Xc/c12/corridor}{
  \path[fill=white] (\X+42,284) rectangle (\X+118,300);
  \sxcamhead{\X+66}{292}{\cam}{\place}}
\cwtab{0}{196}{14}{280}{Grey}{\rotatebox{90}{\scriptsize View}}
\foreach \X/\cam in {\Xa/c03,\Xb/c08,\Xc/c12}{
  \sxcard{p\cam}{\X}{280}{\CW}{84.4}{mtp_\cam.jpg}
  \sxtagbox{p\cam}{mtp_\cam-A}{cwAd}{113}{n}}
\def\Ya{153}\def\Yc{79}
\path[fill=cwA!13,rounded corners=4pt] (\XL-4,\Ya-35) rectangle (\W,\Ya+35);
\path[fill=cwC!18,rounded corners=4pt] (\XL-4,\Yc-35) rectangle (\W,\Yc+35);
\cwtab{0}{\Ya-35}{14}{\Ya+35}{Sky}{\rotatebox{90}{\scriptsize GT 113}}
\cwtab{0}{\Yc-35}{14}{\Yc+35}{Lav}{\rotatebox{90}{\scriptsize GT 067}}
\foreach \x in {\Xb-6,\Xc-6}{\draw[sxwarp] (\x,\Yc-38) -- (\x,284);}
\draw[sxyarn=cwAd] (\XL+12,\Ya) -- (\W-8,\Ya);
\draw[sxyarn=cwCd] (\XL+12,\Yc) -- (\W-8,\Yc);
\sxbobbin{\XL+10}{\Ya}{cwAd}\sxbobbin{\XL+10}{\Yc}{cwCd}
\sxknot{\W-6}{\Ya}{cwAd}\sxknot{\W-6}{\Yc}{cwCd}
\def\bw{33}\def\bh{53.2}
\foreach \X/\cam/\fa/\fb/\gap in {\Xa/c03/1370/1433/+63,\Xb/c08/1011/1068/+57,\Xc/c12/2008/5544/+3{,}536}{
  \sxbead{a\cam1}{\X+44}{\Ya+2}{\bw}{\bh}{mt_113_\cam_\fa.jpg}{cwAd}
  \sxbead{a\cam2}{\X+112}{\Ya+2}{\bw}{\bh}{mt_113_\cam_\fb.jpg}{cwAd}
  \node[sxgap] at (\X+78,\Ya+9) {\gap\ fr};
  \node[inner sep=0pt,font=\scriptsize,text=cwNote,anchor=north] at (\X+44,\Ya-25.6) {f\fa};
  \node[inner sep=0pt,font=\scriptsize,text=cwNote,anchor=north] at (\X+112,\Ya-25.6) {f\fb};}
\foreach \X/\cam/\fa/\fb/\gap in {\Xa/c03/3609/3679/+70,\Xb/c08/3291/3337/+46,\Xc/c12/4195/6406/+2{,}211}{
  \sxbead{c\cam1}{\X+44}{\Yc+2}{\bw}{\bh}{mt_067_\cam_\fa.jpg}{cwCd}
  \sxbead{c\cam2}{\X+112}{\Yc+2}{\bw}{\bh}{mt_067_\cam_\fb.jpg}{cwCd}
  \node[sxgap] at (\X+78,\Yc+9) {\gap\ fr};
  \node[inner sep=0pt,font=\scriptsize,text=cwNote,anchor=north] at (\X+44,\Yc-25.6) {f\fa};
  \node[inner sep=0pt,font=\scriptsize,text=cwNote,anchor=north] at (\X+112,\Yc-25.6) {f\fb};}
\foreach \cam/\k in {c03/2,c08/2,c12/1}{
  \draw[cwAd!70,line width=0.6pt,dash pattern=on 1.6pt off 1.4pt] (p\cam.south -| a\cam\k.north) -- (a\cam\k.north);
  \fill[cwAd] (p\cam.south -| a\cam\k.north) circle[radius=1.3];}
\foreach \Y/\cam/\p in {\Ya/c12/a,\Yc/c12/c}{
  \path[draw=cwHiD,line width=0.8pt,rounded corners=3pt,dash pattern=on 2.4pt off 1.6pt] (\Xc+21,\Y-34) rectangle (\Xc+135,\Y+32);}
\end{tikzpicture}}
\caption{\textbf{Cross-camera continuity across three MTMMC views.} Top: one frame per camera from sequence s16 showing GT~113 (person A of main-paper Figures~1--2 and Figure~\ref{fig:mechanism}) boxed in teal. Lanes: two ground-truth crops per camera for GT~113 and GT~067~\cite{woo2024mtmmc} with camera-local frame indices and inter-observation frame gaps (dashed gold marks the longest gap). Each yarn tracks one true identity despite changes in pose, scale, and occlusion by railings and nearby pedestrians.}
\label{fig:dataset-mtmmc}
\end{figure*}

Figure~\ref{fig:dataset-mtmmc} follows two annotated identities across three cameras of sequence s16, threading each person's ground-truth crops along one continuous yarn. Across stairwell, corridor, and landing views, viewpoint, scale, railing occlusion, and the gap between observations all change from camera to camera.

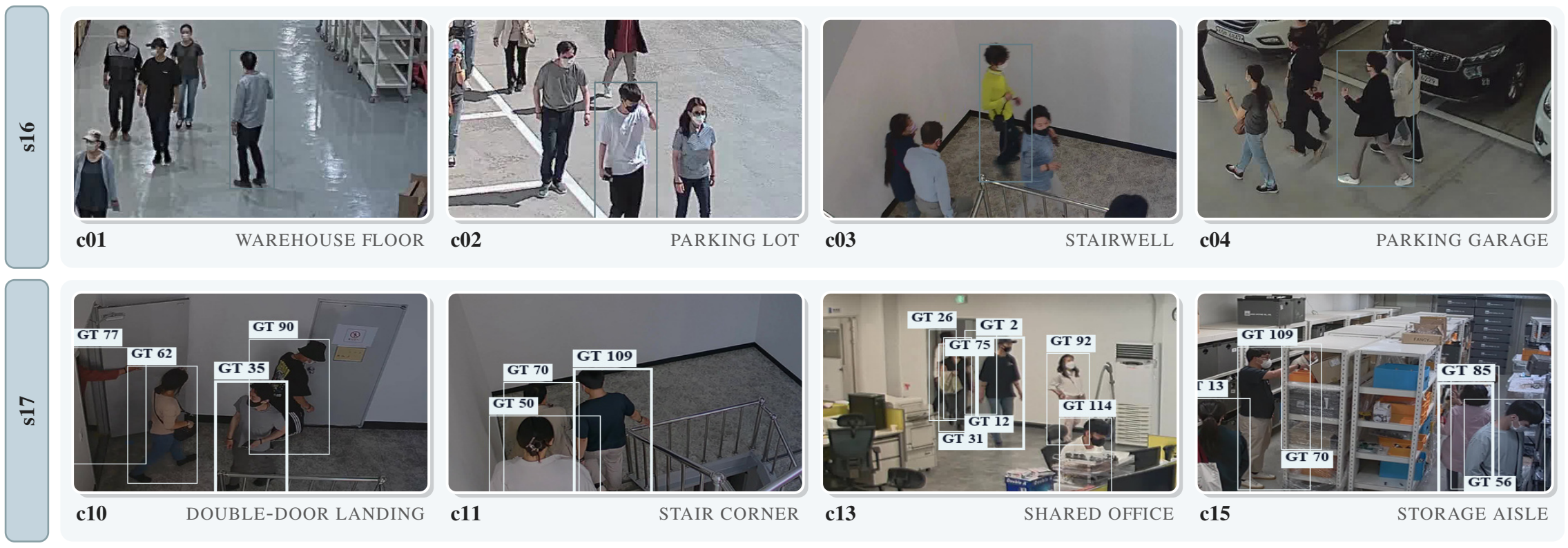
\begin{figure*}[t]\centering
\resizebox{0.86\linewidth}{!}{\begin{tikzpicture}[cw,font=\footnotesize]
\def\gw{117}\def\gh{65.8}\def\gx{22}\def\gs{6}
\foreach \seq/\first/\cams/\places [count=\r from 0] in
  {s16/1/{c01,c02,c03,c04}/{warehouse floor,parking lot,stairwell,parking garage},
   s17/5/{c10,c11,c13,c15}/{double-door landing,stair corner,shared office,storage aisle}}{
  \pgfmathsetmacro{\yt}{-\r*(\gh+24)}
  \path[fill=cwTGrey,rounded corners=4pt] (\gx-4,\yt-\gh-16) rectangle (\gx+4*\gw+3*\gs+4,\yt+4);
  \cwtab{0}{\yt-\gh-16}{14}{\yt+4}{Grey}{\rotatebox{90}{\scriptsize \seq}}
  \foreach \c [count=\i from 0] in \cams {
    \pgfmathtruncatemacro{\k}{\first+\i}
    \pgfmathsetmacro{\xl}{\gx+\i*(\gw+\gs)}
    \edef\gfile{sx/mtg_img0\k.png}
    \sxcardg{\xl}{\yt}{\gw}{\gh}{\gfile}
    \node[anchor=north west,inner sep=0pt,font=\scriptsize\bfseries] at (\xl+1,\yt-\gh-4) {\c};}
  \foreach \p [count=\i from 0] in \places {
    \pgfmathsetmacro{\xl}{\gx+\i*(\gw+\gs)}
    \node[anchor=north east,inner sep=0pt,sc] at (\xl+\gw-1,\yt-\gh-4.6) {\p};}}
\end{tikzpicture}}
\caption{\textbf{Scene and viewpoint diversity across eight MTMMC cameras.} One frame per camera with ground-truth boxes~\cite{woo2024mtmmc}, highlighting a single annotated identity across four views of s16 and all annotated identities with their global labels across four views of s17.}
\label{fig:mtmmc-scenes}
\end{figure*}

Beyond sequence s16, MTMMC spans warehouses, parking areas, offices, and storage aisles where shelving, vehicles, and desks partially occlude pedestrians while co-walking groups enter cameras side by side. Figure~\ref{fig:mtmmc-scenes} illustrates eight camera views from sequences s16 and s17.

\section{Shared groups and scale changes in WILDTRACK}
\label{app:dataset-wildtrack}
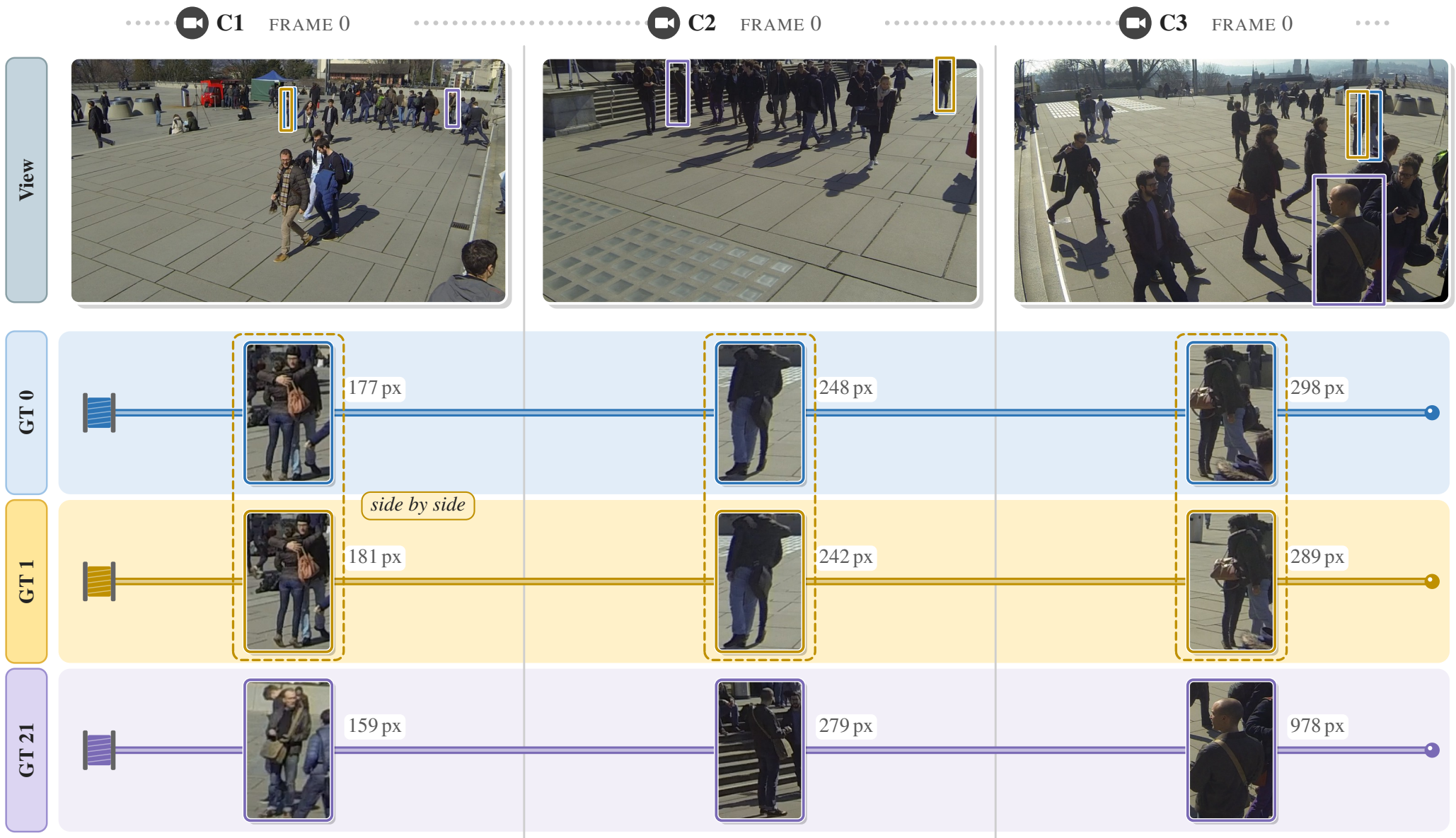
\begin{figure*}[t]\centering
\resizebox{0.78\linewidth}{!}{\begin{tikzpicture}[cw,font=\footnotesize]
\def\W{496}
\path (0,10) (\W,300);
\def\CW{150}\def\XL{22}
\def\Xa{22}\def\Xb{184}\def\Xc{346}
\definecolor{wtA}{HTML}{2E75B6}\colorlet{wtB}{cwHiD}\colorlet{wtC}{cwCd}
\draw[cwInk!25,line width=1.6pt,line cap=round,dash pattern=on 0pt off 3.2pt] (\Xa+20,292) -- (\Xc+130,292);
\foreach \X/\cam in {\Xa/C1,\Xb/C2,\Xc/C3}{
  \path[fill=white] (\X+42,284) rectangle (\X+118,300);
  \sxcamhead{\X+66}{292}{\cam}{frame 0}}
\cwtab{0}{196}{14}{280}{Grey}{\rotatebox{90}{\scriptsize View}}
\foreach \X/\cam in {\Xa/C1,\Xb/C2,\Xc/C3}{
  \sxcard{p\cam}{\X}{280}{\CW}{84.4}{wtp_\cam.jpg}
  \sxgt{p\cam}{wtp_\cam-0}{white,line width=1.7pt}\sxgt{p\cam}{wtp_\cam-0}{wtA,line width=0.9pt}
  \sxgt{p\cam}{wtp_\cam-1}{white,line width=1.7pt}\sxgt{p\cam}{wtp_\cam-1}{wtB,line width=0.9pt}
  \sxgt{p\cam}{wtp_\cam-21}{white,line width=1.7pt}\sxgt{p\cam}{wtp_\cam-21}{wtC,line width=0.9pt}}
\def\Ya{158}\def\Yb{100}\def\Yc{42}
\foreach \Y/\c/\t/\id in {\Ya/cwLSky!30/Sky/0,\Yb/cwGoldL/Amber/1,\Yc/cwC!18/Lav/21}{
  \path[fill=\c,rounded corners=4pt] (\XL-4,\Y-28) rectangle (\W,\Y+28);
  \cwtab{0}{\Y-28}{14}{\Y+28}{\t}{\rotatebox{90}{\scriptsize GT \id}}}
\foreach \x in {\Xb-6,\Xc-6}{\draw[sxwarp] (\x,\Yc-31) -- (\x,284);}
\foreach \Y/\c in {\Ya/wtA,\Yb/wtB,\Yc/wtC}{
  \draw[sxyarn=\c] (\XL+12,\Y) -- (\W-8,\Y);
  \sxbobbin{\XL+10}{\Y}{\c}\sxknot{\W-6}{\Y}{\c}}
\def\bw{30}\def\bh{48.4}
\foreach \Y/\c/\id in {\Ya/wtA/0,\Yb/wtB/1,\Yc/wtC/21}{
  \foreach \X/\cam in {\Xa/C1,\Xb/C2,\Xc/C3}{
    \sxbead{b\id\cam}{\X+75}{\Y}{\bw}{\bh}{wt_\id_\cam.jpg}{\c}
    \node[sxgap,anchor=west] at (\X+94,\Y+8) {\csname sxh@wt-\id-\cam\endcsname\,px};}}
\foreach \X in {\Xa,\Xb,\Xc}{
  \path[draw=cwHiD,line width=0.8pt,rounded corners=3pt,dash pattern=on 2.4pt off 1.6pt] (\X+56,\Yb-27) rectangle (\X+94,\Ya+27);}
\node[pill,anchor=west] at (\Xa+100,{(\Ya+\Yb)/2-3}) {side by side};
\end{tikzpicture}}
\caption{\textbf{Shared group context and cross-camera scale shifts in WILDTRACK.} Top: frame 0 from overlapping cameras C1, C2, and C3 with ground-truth boxes for three pedestrians~\cite{chavdarova2018wildtrack}. Lanes: per-view crops and bounding-box heights in pixels. GT~0 and GT~1 (dashed gold) overlap in every view, while a single person spans 159 to 978 pixels in height across cameras.}
\label{fig:dataset-wildtrack}
\end{figure*}

Figure~\ref{fig:dataset-wildtrack} shows how overlapping outdoor views in WILDTRACK can still leave an arrival ambiguous. At one synchronized instant across cameras C1, C2, and C3, GT~0 and GT~1 walk side by side while GT~21 walks alone, so shared proximity supports both pairings of the co-walking pair equally while large cross-view scale changes strain appearance descriptors.

\section{Indoor viewpoints and occlusions in MMPTrack}
\label{app:dataset-mmptrack}
\begin{figure*}[t]
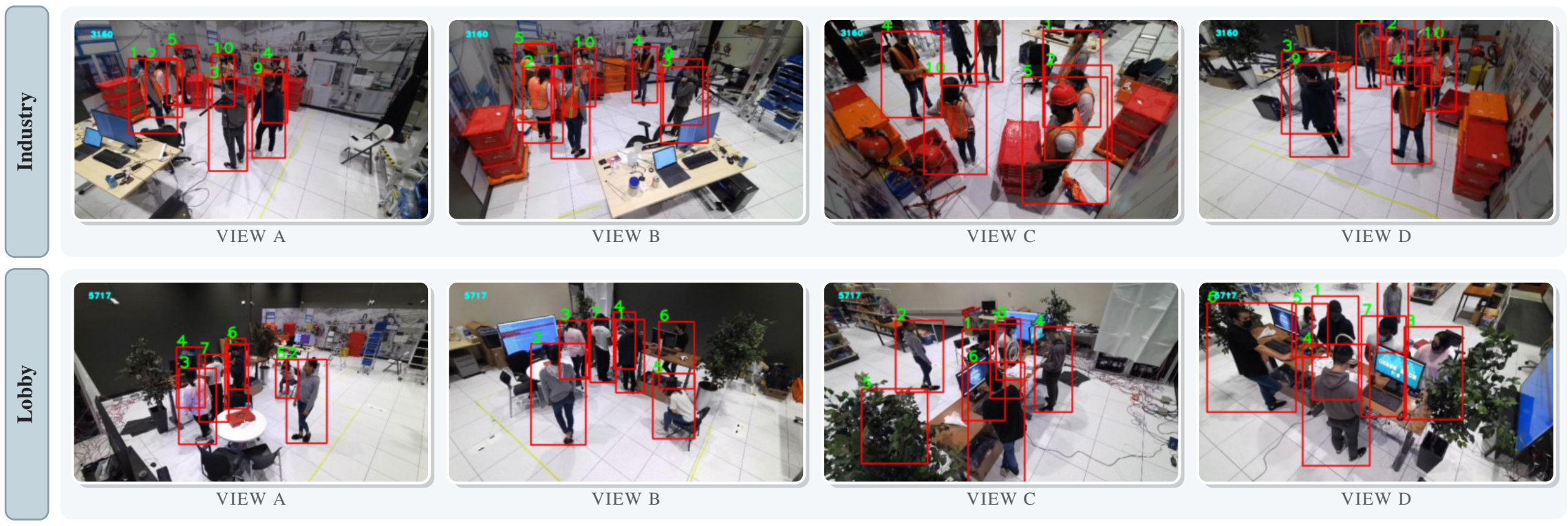
\centering
\resizebox{0.86\linewidth}{!}{\sxgallery{Dmmp}{2}{4}{117}{66.2}{Industry/1,Lobby/5}{a,b,c,d}}
\caption{\textbf{Indoor clutter and cross-view occlusion in MMPTrack.} Four synchronized camera views each from Industry and Lobby, reproduced from Figure~2 of Han et al.~\cite{han2023mmptrack} with original ground-truth boxes and identity labels following the source panel order.}
\label{fig:dataset-mmptrack}
\end{figure*}

In the Industry and Lobby environments of MMPTrack shown in Figure~\ref{fig:dataset-mmptrack}, machinery, furniture, and tight indoor corridors repeatedly occlude body parts and alter apparent pose across views.

\section{Viewpoint and scene variation in DIVOTrack}
\label{app:dataset-divotrack}
\begin{figure*}[t]
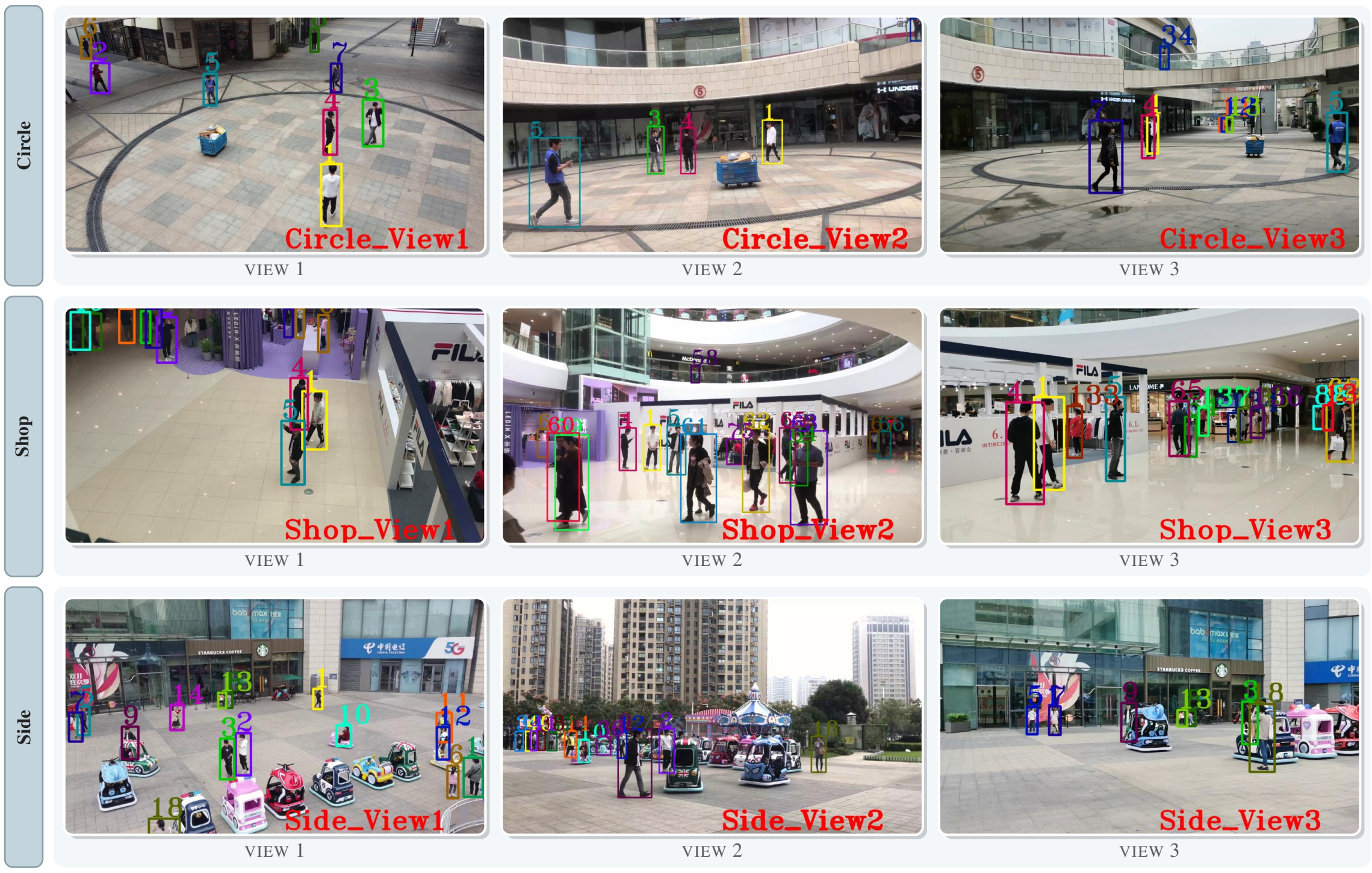
\centering
\resizebox{0.86\linewidth}{!}{\sxgallery{Ddivo}{3}{3}{156}{87.7}{Circle/1,Shop/4,Side/7}{1,2,3}}
\caption{\textbf{Elevation, scale, and background variation across DIVOTrack scenes.} Three synchronized views each from Circle, Shop, and Side, reproduced from Figure~1 of Hao et al.~\cite{hao2024divotrack} with original bounding boxes, identity labels, and camera identifiers. The images are from the DIVOTrack dataset collected by Zhejiang University.}
\label{fig:dataset-divotrack}
\end{figure*}

Figure~\ref{fig:dataset-divotrack} shows the Circle, Shop, and Side environments of DIVOTrack, spanning open pedestrian plazas and a multi-camera indoor mall whose views differ in camera elevation, person scale, crowd overlap, and background.

{\small\bibliographystyle{ieeenat_fullname}\bibliography{supp_references}}